\documentclass{IEEEtaes}

\usepackage{color, array, amsthm}
\usepackage{graphicx}
\usepackage{amsmath, amssymb, amsfonts}
\usepackage{bm}
\usepackage{algorithm}
\usepackage{algorithmic}
\usepackage{xcolor}
\usepackage{comment}
\usepackage{epstopdf}
\usepackage{graphicx}
\usepackage{threeparttable}

\newcommand{\lx}[1]{\textcolor{black}{#1}}
\newcommand{\yjy}[1]{\textcolor{black}{#1}}

\jvol{XX}
\jnum{XX}
\jmonth{XXXXX}
\paper{1234567}
\pubyear{2026}
\doiinfo{TAES.2026.Doi Number}

\newtheorem{theorem}{Theorem}

\newtheorem{proposition}{Proposition}
\newtheorem{remark}{Remark}
\newtheorem{assumption}{Assumption}
\def \bx {\bm{x}}
\def \bX {\bm{X}}
\def \K {\bm{K}}
\def \p {\partial}
\def \N {\mathcal{N}}
\def \bk {\bm{k}}
\def \bK {\bm{K}}

\begin{document}

\title{The Multivariate Feedback Particle Filter for \lx{State-dependent} Multiplicative Noises and Its Application to Attitude and  \lx{Pose Estimations}}

\author{JINGYAO YOU}
\affil{Beihang University, Beijing, China}


\author{XUE LUO}
\member{Senior member, IEEE}
\affil{Beihang University, Beijing, China}


\receiveddate{Manuscript received XXXXX 00, 0000; revised XXXXX 00, 0000; accepted XXXXX 00, 0000.\\
This work is financially supported by National Natural Science Foundation of China (Grant No. 12271019) and the National Key R\&D Program of China (Grant No. 2022YFA1005103).}

\authoraddress{Authors' address: J. You is with School of Mathematical Sciences, Beihang University, Beijing 102206, China (e-mail: \href{mailto:jingyao_you@163.com}{jingyaoyou@buaa.edu.cn}). X. Luo is with School of Mathematical Sciences and Key Laboratory of Mathematics, Informatics and Behavioral Semantics (LMIB), Beihang University, Beijing 100191, China (e-mail: \href{mailto:xluo@buaa.edu.cn}{xluo@buaa.edu.cn}).}

\corresp{{\itshape(Corresponding author: X. Luo).}}


\markboth{YOU ET AL.}{MULTIVARIATE FEEDBACK PARTICLE FILTER FOR MULTIPLICATIVE NOISE}

\maketitle

\begin{abstract}
\lx{Nonlinear filtering with state-dependent multiplicative measurement noise poses a notable challenge for spacecraft attitude determination and robotic pose estimation. This work develops the FPF-DM, a decomposition scheme for feedback particle filters tailored to this family of filtering problems. Extending prior decomposition techniques for additive noise and polynomial measurement models, the proposed method resolves the key difficulty that state-dependent multiplicative noise breaks the polynomial assumption and prevents Hermite spectral backward recursion for the weighted Poisson’s equation. By measure transformation and importance reweighting, we incorporate the inverse measurement covariance into particle weights and rescale the feedback gain to restore the polynomial-form Poisson’s equation, with the decomposition generalized to non-uniformly weighted mixtures for closed-form gain computation. This general FPF-DM is instantiated on quaternion and dual-quaternion manifolds for attitude tracking and point-cloud registration. Simulations on 4D attitude tracking and 8D point-cloud registration verify that the instantiated filters using only 10 particles achieve superior accuracy and computational performance relative to baseline filters.}
\end{abstract}


\section{INTRODUCTION}\label{sec-1}

In the modeling of three-dimensional (3D) attitude tracking on $SO(3)$ \cite{choukroun2023quaternion} and six-degree-of-freedom (6-DoF) pose estimation on $SE(3)$ represented by unit quaternions and dual quaternions \cite{zivan2022dual}, state-dependent multiplicative measurement noise frequently arises. Specifically, rearranging nonlinear geometric constraints into linear pseudo-measurements couples the sensor noise multiplicatively with the state \cite{choukroun2006novel, srivatsan2016estimating}, while practical sensors also exhibit state-dependent noise variances during maneuvers or near point-cloud boundaries \cite{li2022pose}.
Nonlinear filtering (NLF) is a powerful technique to estimate the state of a stochastic dynamical system from noisy observations, \yjy{which plays a key role in spacecraft attitude determination \cite{markley2014fundamentals, crassidis2007survey}, satellite relative navigation \cite{zivan2022dual}, and LiDAR point-cloud registration \cite{srivatsan2016estimating, li2022pose}. }

Classical nonlinear filters such as the extended Kalman filter (EKF) rely on local linearization and often suffer from estimation bias or divergence under strong nonlinearities and large initial errors \cite{crassidis2007survey, sarkka2023bayesian}. Particle filters (PFs) represent non-Gaussian distributions with weighted random samples \cite{arulampalam2002tutorial,gordon1993novel} but suffer from weight degeneracy in high dimensions, requiring large particle ensembles and expensive resampling \cite{surace2019avoid}. Existing quaternion and dual-quaternion filters are mostly EKF or PF variants that inherit the above drawbacks; they either employ linear Kalman updates \cite{zivan2022dual,choukroun2006novel, srivatsan2016estimating} or depend on particle resampling \cite{sveier2021dual}.

\lx{In NLF, Yang et al. \cite{yang2013feedback, yang2016multivariable} developed the feedback particle filter (FPF) rooted in mean-field game theory, which replaces the sequential resampling step of standard PF with a closed-loop feedback control law. The feedback gain of the FPF obeys a weighted Poisson's equation. As numerically validated in \cite{surace2019avoid}, the FPF attains reduced estimation variance and improved scalability relative to conventional PF, provided the gain function can be evaluated accurately. A major bottleneck is that this Poisson's equation generally admits no closed-form solution, motivating a suite of approximation strategies such as constant-gain approximation \cite{yang2016multivariable}, Galerkin projection \cite{yang2013feedback}, proper orthogonal decomposition, and kernel-based gain construction \cite{taghvaei2016gain}; a comparative evaluation of these approximation schemes is reported in \cite{berntorp2018comparison}. For pose estimation, the constant-gain FPF is extended to a dual-quaternion formulation in \cite{li2022pose}. However, it neglects the spatial dependence of the feedback gain and thus suffers from performance degradation for high-order nonlinear measurement models.}

\lx{Very recently, Wang et al. \cite{WML:25, WL:25} developed a decomposition method to derive closed-form gain functions for the FPF under the polynomial measurement assumption, where the underlying conditional density is approximated by an equally weighted particle-based Gaussian mixture (GM). Specifically, the approach decomposes the Poisson's equation into decoupled sub-equations for each particle, which are solved via tensor-product Hermite backward recursion and radial integration. This yields linear computational complexity scaling with the number of particles. Nevertheless, the decomposition framework is limited to systems with additive observation noise and uniformly weighted particles.}

\lx{In this paper, we develop an efficient decomposition scheme for the multivariate FPF to address nonlinear systems with state-dependent multiplicative measurement noise. The primary obstacle to extend the existing decomposition method arises because the state-dependent inverse measurement noise covariance \(\bm{R}^{-1}(\bm{x})\) enters the Poisson’s equation. This term violates the polynomial assumption adopted in prior decomposition frameworks  \cite{WL:25}  and renders the Hermite backward recursion infeasible. To tackle this difficulty, we resort to a measure transformation and importance reweighting strategy, which absorbs the troublesome state-dependent noise covariance term into particle weights and rescales the feedback gain. This treatment restores the polynomial structure required by the Hermite recursion and supports weighted Gaussian mixture approximations. Furthermore, we extend the decomposition approach in \cite{WL:25} to accommodate non-uniformly weighted Gaussian mixtures, which yields a closed-form feedback gain solution via tensor-product Hermite backward recursion and incomplete Gamma functions; the main results are summarized in Proposition~\ref{prop:weighted_decomp}} \yjy{and Theorem~\ref{thm:main_gain}.}

\lx{Built on this general FPF-DM framework, we further construct quaternion and dual-quaternion FPF variants tailored for attitude and pose estimations, termed QFPF-DM and DQFPF-DM, respectively. The proposed filters are validated on two test cases: a 4-dimensional (4D) gyro-aided quaternion attitude tracking problem and  an 8-dimensional (8D) dual-quaternion point-cloud registration problem.  Simulation results demonstrate that the proposed filter, with merely \(N_p = 10\) particles, attains superior estimation accuracy compared with baseline filters, including the KF, the FPF with constant gain (FPF-CG) using 50 particles, and the standard PF with 200 particles,} \yjy{while requiring substantially lower computational overhead than the standard PF.}

The remainder of this paper is organized as follows. Section~\ref{sec-2} briefly formulates the quaternion attitude and dual-quaternion pose estimation problems and reviews the multivariate FPF framework along with the closed-form solutions to the decoupled sub-equations under additive noise. Section~\ref{sec-3} develops the measure transformation, importance reweighting, and generalized weighted decomposition for the FPF. Section~\ref{sec-3.3} instantiates the QFPF-DM and DQFPF-DM to tackle the attitude and pose estimations. Section~\ref{sec-4} presents numerical simulations for 4D attitude tracking and 8D point-cloud registration, and Section~\ref{sec-5} concludes the paper. 

\section{PRELIMINARIES}\label{sec-2}

\yjy{We shall first introduce the quaternion-based attitude and dual-quaternion-based pose estimation models in aerospace and robotic applications in Section \ref{sec-2.1}. Then,} \lx{we review the multivariate FPF \cite{yang2013feedback,yang2016multivariable} for the nonlinear system with additive measurement noise, and the closed-form expression of the gain function obtained by the decomposition method in \cite{WML:25, WL:25} under the polynomial measurement assumption.} Throughout this paper, Cartesian vectors are denoted in bold lowercase font such as $\bm{a}$, quaternions are denoted by a tilde over bold font such as $\tilde{\bm{q}} \in \mathbb{R}^4$, and dual quaternions are denoted by a hat over bold font such as $\hat{\bm{q}} \in \mathbb{R}^8$.

\subsection{\lx{The Models in Attitude and Pose Estimations}}\label{sec-2.1}
\yjy{In aerospace engineering and autonomous navigation---such as spacecraft attitude determination, satellite relative navigation, and LiDAR-based pose estimation---unit quaternions and unit dual quaternions are widely employed to represent 3D rotations on $SO(3)$ and 6-DoF rigid-body poses on $SE(3)$ \cite{choukroun2023quaternion, zivan2022dual}. }\lx{For self-containedness, the definitions and basic properties of unit quaternions and unit dual quaternions are given in the appendix.}

\subsubsection{4D Quaternion Gyroscope Attitude Tracking Model}
\label{sec:model_4d}
Consider a rigid spacecraft or aerial vehicle rotating in 3D space, whose attitude at time $t$ is parameterized by the unit quaternion $\tilde{\bm{q}}_t = [q_{0,t}, \, \bm{q}_{v,t}^T]^T = [q_{0,t}, \, q_{1,t}, \, q_{2,t}, \, q_{3,t}]^T \in \mathcal{S}^3 \subset \mathbb{R}^4$ \cite{markley2014fundamentals}. The body-frame angular velocity is measured by a triaxial gyroscope corrupted by zero-mean Gaussian white noise, \begin{equation}\label{eqn-2.3}
\bm{\omega}_{\mathrm{meas},t} = \bm{\omega}_{\mathrm{true},t} + \sigma_g\dot{\bm{B}}_t,
\end{equation}
where $\bm{B}_t \in \mathbb{R}^3$ is a standard Wiener process, $\dot{\bm{B}}_t = \mathrm{d}\bm{B}_t/\mathrm{d}t$ denotes its formal time derivative, and $\sigma_g > 0$ is the gyroscope noise standard deviation. As modeled in \cite{markley2014fundamentals} , substituting  \eqref{eqn-2.3} into the quaternion kinematics and applying \eqref{eq:quat_vec_reorder} gives: 
\begin{equation*}   
\begin{aligned}
\dot{\tilde{\bm{q}}}_t &=\frac{1}{2}\tilde{\bm{q}}_t \odot\tilde{\bm{\omega}}_{\mathrm{true},t}
= \frac{1}{2}\tilde{\bm{q}}_t \odot \big(\tilde{\bm{\omega}}_{\mathrm{meas},t} - \sigma_g\dot{\tilde{\bm{B}}}_t\big) \\
&= \frac{1}{2}\bm{\Omega}(\bm{\omega}_{\mathrm{meas},t})\tilde{\bm{q}}_t - \frac{1}{2}\sigma_g\bm{\Xi}(\tilde{\bm{q}}_t)\dot{\bm{B}}_t,
\end{aligned}
\end{equation*}
with $\odot$ being the multiplication of quaternions \eqref{eq:quat_mult_def}, which directly yields the state equation \cite{choukroun2023quaternion, choukroun2006novel}:
\begin{equation}
\mathrm{d}\tilde{\bm{q}}_t = \frac{1}{2}\bm{\Omega}(\bm{\omega}_{\mathrm{meas},t})\tilde{\bm{q}}_t\,\mathrm{d}t - \frac{1}{2}\sigma_g\bm{\Xi}(\tilde{\bm{q}}_t)\,\mathrm{d}\bm{B}_t, \label{eq:4d_sde_a}
\end{equation}
where the matrices 
\begin{align*}
\bm{\Omega}(\bm{\omega}) := &\begin{bmatrix} 0 & -\bm{\omega}^T \\ \bm{\omega} & -[\bm{\omega}]^\times \end{bmatrix}\in \mathbb{R}^{4\times 4}, \\
\bm{\Xi}(\tilde{\bm{q}}_t) :=& \begin{bmatrix} -\bm{q}_{v,t}^T \\ q_{0,t}\bm{I}_3 + [\bm{q}_{v,t}]^\times \end{bmatrix}\in \mathbb{R}^{4\times 3},
\end{align*}
with $\bm{I}_3$ being the $3\times 3$ identity matrix and $[\cdot]^\times$ denotes the $3\times 3$ skew-symmetric cross-product matrix \eqref{eq:skew_sym_mat}.

Meanwhile, onboard vector attitude sensors measure reference directions projected into the body frame through the \(3\times 3\) direction cosine matrix \(\bm{C}(\tilde{\bm{q}}_t) \in SO(3)\). Since rotating a 3D reference vector \(\bm{b} \in \mathbb{R}^3\) to obtain \(\bm{a} = \bm{C}(\tilde{\bm{q}})\bm{b}\) is equivalently represented by the quaternion product \(\tilde{\bm{a}} = \tilde{\bm{q}} \odot \tilde{\bm{b}} \odot \tilde{\bm{q}}^*\) \cite{choukroun2023quaternion}, the direction cosine matrix is defined as
\begin{equation}
\label{eq:C_qr_def}
\bm{C}(\tilde{\bm{q}}_t) = (q_{0,t}^2 - \bm{q}_{v,t}^T\bm{q}_{v,t})\bm{I}_3 + 2\bm{q}_{v,t}\bm{q}_{v,t}^T + 2q_{0,t}[\bm{q}_{v,t}]^\times.
\end{equation}
Because each entry of $\bm{C}(\tilde{\bm{q}}_t)$ in \eqref{eq:C_qr_def} is a homogeneous quadratic polynomial of $\tilde{\bm{q}}_t$, the vector sensor measurements are formulated as quadratic forms $h_j(\tilde{\bm{q}}_t) = \tilde{\bm{q}}_t^T\bm{S}_j\tilde{\bm{q}}_t$ ($p = 2$) with known symmetric matrices $\bm{S}_j \in \mathbb{R}^{4\times 4}$. 

\yjy{For quaternion attitude determination, sensor degradation from large inclination angles and solar glint introduces state-dependent noise variance \(R_j(\tilde{\bm{q}}_t, t) > 0\). This produces multiplicative measurement noise, with \(\sqrt{R_j(\tilde{\bm{q}}_t, t)}\) scaling the Wiener increment \(\mathrm{d}W_{j,t}\) in the observation equation \cite{choukroun2023quaternion}:}
\begin{align}
\mathrm{d}Z_{j,t} &= \big(\tilde{\bm{q}}_t^T\bm{S}_j\tilde{\bm{q}}_t\big)\,\mathrm{d}t + \sqrt{R_j(\tilde{\bm{q}}_t, t)}\,\mathrm{d}W_{j,t},\label{eq:4d_sde_b}
\end{align}
where $\{W_{j,t}\}_{j=1}^m$ are independent standard Wiener processes.

\lx{The objective of 4D quaternion attitude estimation is to estimate $\tilde{\bm{q}}_t$ in \eqref{eq:4d_sde_a} with the observation history $Z_{j,t}$ upto time $t$.}

\subsubsection{8D Dual Quaternion Pose Estimation Model}
\label{sec:model_8d}
Given a reference 3D point cloud $\{\bm{b}_k\}_{k=1}^{N_{\mathrm{pts}}} \subset \mathbb{R}^3$ and its transformed sensor measurements $\{\bm{a}_k\}_{k=1}^{N_{\mathrm{pts}}} \subset \mathbb{R}^3$ related by the rigid-body transformation $\bm{a}_k = \bm{C}(\tilde{\bm{q}}_r)\bm{b}_k + \bm{t}$, where $\tilde{\bm{q}}_r = [q_0, \, \bm{q}_v^T]^T \in \mathcal{S}^3$ is the unit rotation quaternion, $\bm{t} = [t_x, \, t_y, \, t_z]^T \in \mathbb{R}^3$ is the translation vector, and $\bm{C}(\tilde{\bm{q}}_r)$ is defined in \eqref{eq:C_qr_def}, the objective of point cloud registration is to estimate the unknown static 6-DoF pose parameterized by the 8D unit dual quaternion \cite{zivan2022dual, li2022pose}:
\begin{equation}
\label{eq:unit_dq_se3}
\hat{\bm{q}} = \tilde{\bm{q}}_r + \varepsilon\tilde{\bm{q}}_d=[\tilde{\bm{q}}_r^T,\tilde{\bm{q}}_d^T]^T \in \mathbb{R}^8, 
\end{equation}
with $\tilde{\bm{q}}_d = \frac{1}{2}\tilde{\bm{t}} \odot \tilde{\bm{q}}_r$, where $\varepsilon$ is Clifford's dual unit, $\tilde{\bm{t}} = [0, \, \bm{t}^T]^T \in \mathbb{R}^4$ is the pure translation quaternion associated with $\bm{t}$, and $\tilde{\bm{q}}_d \in \mathbb{R}^4$ is the dual part.
\yjy{Here, the ``$+$'' sign in \eqref{eq:unit_dq_se3} denotes the formal algebraic combination over the dual-number ring rather than standard vector addition in $\mathbb{R}^4$, so $\hat{\bm{q}}$ is represented in vector form by stacking $\tilde{\bm{q}}_r$ and $\tilde{\bm{q}}_d$.}

Since the rigid-body pose is static during scan registration, the filtering system for the 4D rotational state $\tilde{\bm{q}}_{r,t} \in \mathcal{S}^3$ at measurement step $k$ is governed by:
\begin{equation}
\mathrm{d}\tilde{\bm{q}}_{r,t} = \bm{0}.\label{eq:8d_sde_a}
\end{equation}

As derived in the appendix, embedding $\bm{a}_k, \bm{b}_k \in \mathbb{R}^3$ as pure quaternions $\tilde{\bm{a}}_k = [0, \, \bm{a}_k^T]^T$ and $\tilde{\bm{b}}_k = [0, \, \bm{b}_k^T]^T$ and subtracting the dual quaternion transformation constraints for two consecutive point pairs $(\bm{a}_k, \bm{b}_k)$ and $(\bm{a}_{k-1}, \bm{b}_{k-1})$ eliminates the dual part $\tilde{\bm{q}}_d$, yielding a decoupled linear observation constraint $\bm{M}_k\tilde{\bm{q}}_r = \tilde{\bm{0}}$, where 
\begin{equation}
\label{eq:Mk_def}
\bm{M}_k := \bm{L}(\tilde{\bm{a}}_k - \tilde{\bm{a}}_{k-1}) - \bm{R}(\tilde{\bm{b}}_k - \tilde{\bm{b}}_{k-1}),
\end{equation}
with $\bm{L}(\cdot)$ and $\bm{R}(\cdot)$ denoting the $4\times 4$ left- and right-quaternion multiplication matrices defined in \eqref{eq:quat_L_mat}--\eqref{eq:quat_R_mat} \cite{srivatsan2016estimating, li2022pose}. \yjy{Since the noisy sensor measurements $\bm{a}_k$ enter $\bm{M}_k$ in \eqref{eq:Mk_def}, the coordinate measurement error is multiplicatively coupled with the unknown state $\tilde{\bm{q}}_{r,t}$ via $\bm{\Gamma}_k(\tilde{\bm{q}}_{r,t})\,\mathrm{d}\bm{W}_t$, which induces the state-dependent observation noise covariance matrix $\bm{R}_k(\tilde{\bm{q}}_{r,t}):= \bm{\Gamma}_k(\tilde{\bm{q}}_{r,t})\bm{\Gamma}_k^T(\tilde{\bm{q}}_{r,t})>0$ \cite{li2022pose}.} \lx{Thus, the observation equation is modeled as}
\begin{align}
\mathrm{d}\bm{Z}_{k,t} = \bm{M}_k\tilde{\bm{q}}_{r,t}\,\mathrm{d}t + \bm{\Gamma}_k(\tilde{\bm{q}}_{r,t})\,\mathrm{d}\bm{W}_t. \label{eq:8d_sde_b}
\end{align}

 Once the posterior rotation estimate $\tilde{\bm{q}}_{r,\mathrm{est},k}$ is obtained from \eqref{eq:8d_sde_a} and \eqref{eq:8d_sde_b}, the translation estimate $\bm{t}_{\mathrm{est},k}$ is recovered in closed form via:
\begin{equation}
\label{eq:dq_sum_trans_eq}
\bm{t}_{\mathrm{est},k} = \frac{\bm{a}_k + \bm{a}_{k-1}}{2} - \bm{C}(\tilde{\bm{q}}_{r,\mathrm{est},k})\frac{\bm{b}_k + \bm{b}_{k-1}}{2},
\end{equation}
and the dual part is recovered as $\tilde{\bm{q}}_{d,\mathrm{est},k} = \frac{1}{2}\tilde{\bm{t}}_{\mathrm{est},k} \odot \tilde{\bm{q}}_{r,\mathrm{est},k}$ with $\tilde{\bm{t}}_{\mathrm{est},k} = [0, \, \bm{t}_{\mathrm{est},k}^T]^T$.

\subsection{The Multivariate Feedback Particle Filter (FPF)}
\label{sec-2.2}
Let us first consider a classical continuous-time nonlinear filtering (NLF) problem \lx{without state-dependent multiplicative measurement noise:}
\begin{subequations}
\label{eq:std_nlf_sde}
\begin{align}
\mathrm{d}\bm{X}_t &= \bm{g}(\bm{X}_t)\,\mathrm{d}t + \bm{\sigma}(\bm{X}_t)\,\mathrm{d}\bm{B}_t, \label{eq:std_nlf_sde_a} \\
\mathrm{d}\bm{Z}_t &= \bm{h}(\bm{X}_t)\,\mathrm{d}t + \mathrm{d}\bm{W}_t, \label{eq:std_nlf_sde_b}
\end{align}
\end{subequations}
where $\bm{X}_t \in \mathbb{R}^d$ denotes the $d$-dimensional state vector at time $t \ge 0$, and $\bm{Z}_t \in \mathbb{R}^m$ represents the $m$-dimensional observation process. The processes $\{\bm{B}_t\}_{t \ge 0}$ and $\{\bm{W}_t\}_{t \ge 0}$ are mutually independent standard Wiener processes taking values in $\mathbb{R}^d$ and $\mathbb{R}^m$, respectively, both independent of the initial state $\bm{X}_0$. The drift vector field $\bm{g}(\cdot): \mathbb{R}^d \to \mathbb{R}^d$ and diffusion matrix $\bm{\sigma}(\cdot): \mathbb{R}^d \to \mathbb{R}^{d\times d}$ satisfy standard Lipschitz continuity and linear growth conditions, while the observation mapping $\bm{h}(\cdot) = (h_1, \dots, h_m)^T: \mathbb{R}^d \to \mathbb{R}^m$ is bounded and continuously differentiable. The objective of NLF is to recursively approximate the posterior probability density function (PDF) $p_t^*(\bm{x}) = \mathbb{P}(\bm{X}_t \in \mathrm{d}\bm{x} \mid \mathcal{Z}_t)$, where $\mathcal{Z}_t := \sigma\{\bm{Z}_s : 0 \le s \le t\}$ \lx{is the observation history up to time $t$.}

Rather than relying on sequential resampling in PF, which suffers from weight degeneracy in high-dimensional spaces \cite{surace2019avoid}, the FPF \cite{yang2013feedback,yang2016multivariable} constructs an ensemble of $N_p$ interacting particles $\{\bm{X}_t^i\}_{i=1}^{N_p} \subset \mathbb{R}^d$. Each particle $\bm{X}_t^i = (X_{t,1}^i, \dots, X_{t,d}^i)^T$ evolves according to a controlled It\^{o} diffusion system driven by feedback control:
\begin{equation}
\label{eq:fpf_particle_sde}
\begin{aligned}
\mathrm{d}\bm{X}_t^i &= \bm{g}(\bm{X}_t^i)\,\mathrm{d}t + \bm{\sigma}(\bm{X}_t^i)\,\mathrm{d}\bm{B}_t^i \\
&\quad + \underbrace{\bm{u}(\bm{X}_t^i, t)\,\mathrm{d}t + \bm{K}(\bm{X}_t^i, t)\,\mathrm{d}\bm{Z}_t}_{\mathrm{d}\bm{U}_t^i},
\end{aligned}
\end{equation}
for $i = 1, \dots, N_p$, where $\{\bm{B}_t^i\}_{i=1}^{N_p}$ are mutually independent standard Wiener processes replicated for each particle, and $\mathrm{d}\bm{U}_t^i$ represents the feedback control process. Here, $\bm{K}(\bm{x}, t) \in \mathbb{R}^{d\times m}$ is the state-dependent feedback gain function, and
\begin{equation}
\label{eq:fpf_drift_u}
\bm{u}(\bm{x}, t) = -\frac{1}{2}\bm{K}(\bm{x}, t)\left(\bm{h}(\bm{x}) + p_t[\bm{h}_t]\right)+ \bm{\Omega}(\bm{x}, t)\in \mathbb{R}^d
\end{equation}
is the drift control vector field. Here, $p_t[\bm{h}_t] \triangleq \int_{\mathbb{R}^d} \bm{h}(\bm{x})p_t(\bm{x})\,\mathrm{d}\bm{x} \in \mathbb{R}^m$, where $p_t(\bx)$ is the empirical distribution of the controlled particle population  $\{\bX_t^i\}_{i=1}^{N_p}$, and $\bm{\Omega}(\bm{x}, t) = [\Omega_1(\bm{x}, t), \dots, \Omega_d(\bm{x}, t)]^T \in \mathbb{R}^d$ is the Wong-Zakai correction term, with 
\begin{equation}
\label{eq:wong_zakai}
\Omega_l(\bm{x}, t) := \frac{1}{2}\sum_{k=1}^{d}\sum_{s=1}^{m} K_{ks}(\bm{x}, t)\frac{\partial K_{ls}(\bm{x}, t)}{\partial x_k},
\end{equation}
where $K_{ks}(\bm{x}, t)$ denotes the $(k, s)$-th entry of $\bm{K}(\bm{x}, t)$. 

Thus, the key ingredient in deriving the multivariate FPF lies in solving for the feedback gain function $\bm{K}(\bm{x}, t) = [\bm{K}_{\cdot 1}(\bm{x}, t), \dots, \bm{K}_{\cdot m}(\bm{x}, t)] \in \mathbb{R}^{d\times m}$. By minimizing the Kullback-Leibler (K-L) divergence between $p_t(\bx)$ and the true posterior $p_t^*(\bm{x})$ via the calculus of first variations, one obtains the Euler-Lagrange boundary value problem (EL-BVP) satisfied by the gain function:
\begin{equation}
\label{eqn-2.1}
\nabla^T\big(p_t(\bm{x})\K_{\cdot,j}(\bm{x})\big) = -\big(h_j(\bm{x}) - p_t[h_{j,t}]\big)p_t(\bm{x}),
\end{equation}
for $j=1,\dots,m$, where $\nabla^T(\cdot) =\sum_{i=1}^d\frac{\p\cdot}{\p x_i}$ denotes the divergence operator in $\mathbb{R}^d$.

However, in general the solution to \eqref{eqn-2.1} is not unique and has no closed-form. To obtain an approximation, we express $\K$'s $j$-th column as the gradient of a scalar potential function $\varphi_j:\,\mathbb{R}^d \to \mathbb{R}$, i.e. $\bm{K}_{\cdot j}(\bm{x}, t) = \nabla\varphi_j(\bm{x}, t)$, $j = 1, \dots, m$. Then \eqref{eqn-2.1} becomes
\begin{equation}
\label{eq:std_poisson_pde}
\nabla^T\big(p_t(\bm{x})\nabla\varphi_j(\bm{x})\big) = -\big(h_j(\bm{x}) -  p_t[h_{j,t}]\big)p_t(\bm{x}),
\end{equation}
with the vanishing flux boundary condition \cite{WL:25} 
\begin{equation}
\label{eq:std_bc_inf}
\begin{aligned}
\lim_{R \to \infty} \int_{S^d(R)} p_t(\bm{x})\nabla\varphi_j(\bm{x}) \cdot \mathrm{d}\bm{\sigma} &= 0, 
\end{aligned}
\end{equation}
where $S^d(R)$ denotes the surface of a $d$-dimensional sphere of radius $R$. \lx{ By classical partial differential equation theory, the solution to \eqref{eq:std_poisson_pde}--\eqref{eq:std_bc_inf} is well-posed, but still has no closed-form in general. Therefore, the only issue is how to approximate the solution $\nabla\varphi_j$ efficiently and accurately. }

\subsection{\lx{The Decomposition Method for FPF \cite{WL:25}}}
\label{sec-2.3}
\lx{To solve for $\nabla\varphi_j(\bm{x})$ in \eqref{eq:std_poisson_pde}, one needs to approximate $p_t(\bm{x})$. By the standard theory of kernel density estimation in nonparametric statistics, Wang et al. \cite{WL:25} }approximate the unknown density $p_t(\bm{x})$ by an equally weighted GM, i.e.,
\begin{equation}
\label{eq:unweighted_gmm}
p_t^{N_p, \bm{\Sigma}}(\bm{x}) = \frac{1}{N_p}\sum_{i=1}^{N_p} \mathcal{N}(\bm{x}; \bm{X}_t^i, \bm{\Sigma}),
\end{equation}
where each Gaussian kernel $\mathcal{N}(\bm{x}; \bm{X}_t^i, \bm{\Sigma})$ is centered at the particle $\bm{X}_t^i$ with a common positive-definite diagonal bandwidth covariance matrix $\bm{\Sigma} = \mathrm{diag}(\epsilon_1, \dots, \epsilon_d) \in \mathbb{R}^{d\times d}$ ($\epsilon_q > 0$), i.e.,
\begin{equation}
\label{eq:gaussian_kernel_def}
\begin{aligned}
\mathcal{N}(\bm{x}; \bm{X}_t^i, \bm{\Sigma}) &= \frac{1}{(2\pi)^{d/2}|\bm{\Sigma}|^{1/2}} \\
&\quad \cdot \exp\left\{-\frac{1}{2}(\bm{x} - \bm{X}_t^i)^T\bm{\Sigma}^{-1}(\bm{x} - \bm{X}_t^i)\right\}.
\end{aligned}
\end{equation}
Here, $|\bm{\Sigma}| = \prod_{q=1}^{d} \epsilon_q$ denotes the determinant of $\bm{\Sigma}$. \lx{Let us denote $\mathcal{N}_i(\bm{x}) := \mathcal{N}(\bm{x} ; \bm{X}_t^i, \bm{\Sigma})$ in the sequel for brevity. The error between $p_t(\bx)$ and $p_t^{N_p,\Sigma}(\bx)$ has been further studied under some conditions in a later version of \cite{WL:25}.}

\lx{Choosing $\bm{\Sigma}$ as a diagonal matrix is not mandatory. }If a non-diagonal positive-definite bandwidth matrix is preferred, it can always be reduced to the diagonal form via an orthogonal coordinate rotation.

\lx{Replacing $p_t(\bm{x})$ in \eqref{eq:std_poisson_pde} with $p_t^{N_p, \bm{\Sigma}}(\bm{x})$ yields} 
\begin{equation}
\label{eq:reg_poisson_pde}
\nabla^T\big(p_t^{N_p, \bm{\Sigma}}(\bm{x})\nabla\varphi_j(\bm{x})\big) = -\big(h_j(\bm{x}) - p_t^{N_p, \bm{\Sigma}}[h_j]\big)p_t^{N_p, \bm{\Sigma}}(\bm{x}),
\end{equation}
where $p_t^{N_p, \bm{\Sigma}}[h_j] = \int_{\mathbb{R}^d} h_j(\bm{x})p_t^{N_p, \bm{\Sigma}}(\bm{x})\,\mathrm{d}\bm{x}$. Unlike the scalar case ($d = 1$) where \eqref{eq:reg_poisson_pde} can be solved via direct spatial integration, the multivariate divergence operator $\nabla^T$ for $d \ge 2$ prevents \lx{the direct integration}. To overcome this, Wang et al. \cite{WML:25, WL:25} developed a decomposition method that splits \eqref{eq:reg_poisson_pde} into $N_p$ pairs of decoupled sub-equations\lx{, which can be solved explicitly. We summarize the main results below.}
\yjy{\begin{proposition}[Proposition III.1, \cite{WML:25}]
\label{prop:prelim_decomp}
The $j$-th column of the gain function $\bm{K}_{\cdot j}(\bm{x}) = \nabla\varphi_j(\bm{x})$ in \eqref{eq:reg_poisson_pde} \lx{can be expressed explicitly as}
\begin{equation}
\label{eq:unweighted_decomp_recon}
\begin{aligned}
\nabla\varphi_j(\bm{x}) &= \frac{1}{\sum_{i=1}^{N_p} \mathcal{N}_i(\bm{x})} \sum_{i=1}^{N_p} \Big[\nabla^{\bm{\Sigma}}\psi_j^i(\bm{x}) + \mathcal{N}_i(\bm{x})\nabla\varphi_j^i(\bm{x})\Big],
\end{aligned}
\end{equation}
where 
\begin{equation}
\label{eq:inv_eig_grad_op}
    \nabla^{\bm{\Sigma}}\psi_j^i(\bm{x}) = \left(\epsilon_1\frac{\partial\psi_j^i(\bm{x})}{\partial x_1}, \dots, \epsilon_d\frac{\partial\psi_j^i(\bm{x})}{\partial x_d}\right)^T \in \mathbb{R}^d,
\end{equation}
denotes the bandwidth-scaled gradient vector field,
and for each particle $i \in \{1, \dots, N_p\}$ and measurement $j \in \{1, \dots, m\}$, the local potential functions $\varphi_j^i(\bm{x})$ and $\psi_j^i(\bm{x})$ satisfy the two decoupled sub-equations:
\begin{align}
\nabla^T\big(\mathcal{N}_i(\bm{x})\nabla\varphi_j^i(\bm{x})\big) &= \big(\!-h_j(\bm{x}) + C_j^i\big)\mathcal{N}_i(\bm{x}), \label{eq:prelim_sub_pde_1} 
\end{align}
and
\begin{align}
\nabla^T\big(\nabla^{\bm{\Sigma}}\psi_j^i(\bm{x})\big) &= \big(p_t^{N_p, \bm{\Sigma}}[h_j] - C_j^i\big)\mathcal{N}_i(\bm{x}), \label{eq:prelim_sub_pde_2}
\end{align}
respectively, with $C_j^i \in \mathbb{R}$ being a decoupling constant associated with the $i$-th particle.
\end{proposition}
}
\yjy{To solve the first sub-equation \eqref{eq:prelim_sub_pde_1} analytically,} \lx{one assumes that
\begin{assumption}\label{as-1}
    Each observation function $h_j(\bm{x})$ is a polynomial of degree at most $p \in \mathbb{Z}_+$, written as \begin{equation}
\label{eq:hermite_obs_assumption}
h_j(\bm{x}) = \sum_{\bm{k} \in \Omega} a_{j,\bm{k}}H_{\bm{k}}(\bm{x}),
\end{equation}
where $H_{\bm{k}}(\bm{x}) := \prod_{q=1}^{d} H_{k_q}(x_q)$, with $H_n(x) = (-1)^n e^{x^2}\frac{\mathrm{d}^n}{\mathrm{d}x^n}e^{-x^2}$ being the univariate Hermite polynomial of degree $n$, $\Omega := \{\bm{k} = (k_1, \dots, k_d) \in \mathbb{N}^d : 0 \le |\bm{k}|_1 = \sum_{q=1}^{d} k_q \le p\}$ is the multi-index set, and $a_{j,\bm{k}} \in \mathbb{R}$ are expansion coefficients.
\end{assumption}}

 \lx{The following proposition gives a backward recursion to solve \eqref{eq:prelim_sub_pde_1}.}
\yjy{\begin{proposition}[Proposition III.1, \cite{WL:25}]
\label{prop:prelim_sub1}
Under the polynomial assumption \eqref{eq:hermite_obs_assumption}, the first sub-equation \eqref{eq:prelim_sub_pde_1} admits an exact polynomial solution \lx{of the form:}
\begin{equation}
\label{eq:phi_hermite_exp}
\varphi_j^i(\bm{x}) = \sum_{\bm{k} \in \Omega} \tilde{\Phi}_{j,\bm{k}}^i H_{\bm{k}}(\bm{x}),
\end{equation}
where the coefficients $\{\tilde{\Phi}_{j,\bm{k}}^i\}_{\bk\in\Omega}$ are uniquely determined from $|\bm{k}|_1 = p$ down to $|\bm{k}|_1 = 1$ via the backward recursion:
\begin{equation}
\label{eq:backward_recursion}
\begin{aligned}
&\left(\sum_{q=1}^{d} \frac{k_q}{\epsilon_q}\right)\tilde{\Phi}_{j,\bm{k}}^i\\
=& a_{j,\bm{k}} + \sum_{q=1}^{d} \frac{2X_{t,q}^i(k_q + 1)}{\epsilon_q}\tilde{\Phi}_{j,\bm{k}+\bm{e}_q}^i \\
&+ \sum_{q=1}^{d} 2(k_q + 2)(k_q + 1)\left(2 - \frac{1}{\epsilon_q}\right)\tilde{\Phi}_{j,\bm{k}+2\bm{e}_q}^i,
\end{aligned}
\end{equation}
with the initial conditions $\tilde{\Phi}_{j,\bm{k}}^i \equiv 0$ for $|\bm{k}|_1 > p$ and terminating with any index $k_q < 0$. Furthermore, the decoupling constant $C_j^i$ is uniquely determined at $\bm{k} = \bm{0}$ as:
\begin{equation}
\label{eq:Ci_formula}
C_j^i = a_{j,\bm{0}} + \sum_{q=1}^{d} \frac{2X_{t,q}^i}{\epsilon_q}\tilde{\Phi}_{j,\bm{e}_q}^i + 4\sum_{q=1}^{d} \left(2 - \frac{1}{\epsilon_q}\right)\tilde{\Phi}_{j,2\bm{e}_q}^i,
\end{equation}
 where $\bm{e}_q \in \mathbb{N}^d$ denotes the $q$-th unit basis vector. Consequently, $p_t^{N_p, \bm{\Sigma}}[h_j] = \frac{1}{N_p}\sum_{i=1}^{N_p} C_j^i$ and  the gradient is given explicitly by $\nabla\varphi_j^i(\bm{x}) = \sum_{\bm{k} \in \Omega} \tilde{\Phi}_{j,\bm{k}}^i \big[2k_1 H_{\bm{k}-\bm{e}_1}(\bm{x}), \dots, 2k_d H_{\bm{k}-\bm{e}_d}(\bm{x})\big]^T$.
\end{proposition}
}
\lx{As for \eqref{eq:prelim_sub_pde_2}, a weighted radially symmetric solution is obtained.}
\yjy{\begin{proposition}[Proposition III.3, \cite{WL:25}]
\label{prop:prelim_sub2}
With $C_j^i$ in \eqref{eq:Ci_formula}, let 
\begin{align}
\label{eq:radial_dist_r}\notag
r_i(\bx) :=& \sqrt{\sum_{q=1}^{d} \frac{(x_q - X_{t,q}^i)^2}{\epsilon_q}}\\
    =& \sqrt{(\bm{x} - \bm{X}_t^i)^T\bm{\Sigma}^{-1}(\bm{x} - \bm{X}_t^i)}
\end{align}
denote the dimensionless weighted radial distance centered at particle $\bm{X}_t^i$. Under the regularity condition at the origin $\nabla^{\bm{\Sigma}}\psi_j^i(\bm{X}_t^i) = \bm{0}$, the second sub-equation \eqref{eq:prelim_sub_pde_2} admits an analytical radially symmetric solution:
\begin{equation}
\label{eq:psi_grad_unweighted}
    \nabla^{\bm{\Sigma}}\psi_j^i(\bm{x}) = (\bm{x} - \bm{X}_t^i)\frac{p_t^{N_p, \bm{\Sigma}}[h_j] - C_j^i}{2\pi^{d/2}|\bm{\Sigma}|^{1/2}}\gamma\left(\frac{d}{2}, \frac{r_i^2}{2}\right)r_i^{-d},
\end{equation}
where $\gamma(s, u) := \int_0^u t^{s-1}e^{-t}\,\mathrm{d}t$ is the lower incomplete Gamma function.
\end{proposition}
}

\yjy{Despite its high accuracy and $\mathcal{O}(N_p)$ computational complexity, the above decomposition framework relies on two restrictive assumptions: 1) the observation noise covariance in \eqref{eq:std_nlf_sde_b} is state-independent, and 2) all particles in the GM \eqref{eq:unweighted_gmm} carry uniform weights $1/N_p$. Section~\ref{sec-3} extends this framework to nonlinear systems with state-dependent multiplicative measurement noise.}

\section{MULTIVARIATE FEEDBACK PARTICLE FILTER FOR THE MULTIPLICATIVE NOISE}\label{sec-3}

\lx{As outlined in Section \ref{sec-2.1}, we first cast the problem into a general NLF formulation with state-dependent multiplicative measurement noise. An FPF decomposition method is then developed via importance reweighting, accompanied by a proof of the decomposition theorem for weighted Gaussian mixtures. This adapted decomposition is further specialized to quaternion and dual-quaternion settings to obtain QFPF-DM and DQFPF-DM for the target estimation tasks.}

\subsection{\lx{The General NLF Model and the FPF}}

\yjy{Both the 4D attitude tracking system \eqref{eq:4d_sde_a} and \eqref{eq:4d_sde_b} and the 8D pose registration system \eqref{eq:8d_sde_a} and \eqref{eq:8d_sde_b} in Section~\ref{sec-2.1} can be abstracted into the general $d$-dimensional NLF problem \lx{with the state-dependent multiplicative noise \cite{yu2022adaptive}}, whose state is governed by the It\^{o} SDE \eqref{eq:std_nlf_sde_a}, while the measurement is}  
\begin{equation}
\label{eq:mult_nlf_sde}
\mathrm{d}\bm{Z}_t = \bm{h}(\bm{X}_t)\,\mathrm{d}t + \bm{\Gamma}(\bm{X}_t)\,\mathrm{d}\bm{W}_t,
\end{equation}
where $\bm{X}_t \in \mathbb{R}^d$ is the state vector, $\bm{Z}_t \in \mathbb{R}^m$ is the observation process, and $\{\bm{B}_t\}, \{\bm{W}_t\}$ are mutually independent standard Wiener processes taking values in $\mathbb{R}^d$ and $\mathbb{R}^m$, respectively. \lx{Besides Assumption \ref{as-1}, we further assume that 
\begin{assumption}\label{as-2}
    $\bm{R}(\bm{x}) := \bm{\Gamma}(\bm{x})\bm{\Gamma}^T(\bm{x}) \in \mathbb{R}^{m\times m}$ is a strictly positive-definite diagonal matrix, written in the form $\bm{R}(\bm{x}) = \mathrm{diag}(R_1(\bm{x}), \dots, R_m(\bm{x}))$.
\end{assumption}}

The state-dependent measurement noise covariance $\bm{R}(\bm{x})$ modifies the EL-BVP \eqref{eqn-2.1} governing the feedback gain function $\bm{K}(\bm{x}) = [\bm{K}_{\cdot 1}(\bm{x}), \dots, \bm{K}_{\cdot m}(\bm{x})] \in \mathbb{R}^{d\times m}$. Specifically, for each measurement  $j \in \{1, \dots, m\}$, the $j$-th column of the gain function $\bm{K}_{\cdot j}(\bm{x}) \in \mathbb{R}^d$ satisfies the heterogeneous Poisson's equation \lx{\cite{li2022pose}:
\begin{equation}
\label{eq:hetero_poisson}
\nabla^T\big(p_t(\bm{x})\bm{K}_{\cdot j}(\bm{x})\big) = -\big(h_j(\bm{x}) - \bar h_j\big)R_j^{-1}(\bm{x})p_t(\bm{x}),
\end{equation}
for some constant $\bar h_j$.} Attempting to apply the decomposition method in Section~\ref{sec-2.3} directly to \eqref{eq:hetero_poisson} \lx{ is infeasible, since the product $\big(h_j(\bm{x}) - \bar h_j\big)R_j^{-1}(\bm{x})$ by no means is a polynomial in $\bx$, even if $h_j(\bm{x})$ is by Assumption \ref{as-1}. }Consequently, the Hermite spectral expansion \eqref{eq:phi_hermite_exp} and the backward recursion \eqref{eq:backward_recursion} cannot be applied directly.

\subsection{Importance-Reweighted GM Approximation and \lx{Weighted Decomposition Method}}
\label{sec:measure_trans}

\lx{To solve \eqref{eq:hetero_poisson} for $\bK_{\cdot j}$, we} introduce a measure transformation that decouples the state-dependent measurement noise variance $R_j(\bx)$ from the polynomial observation function. For each observation $j \in \{1, \dots, m\}$, we define a new probability density 
\begin{equation}\label{eq:tilde_p_def}
    \begin{aligned}
   \tilde{p}_{t,j}(\bm{x}) :=& \frac{1} {C_{R,j}}R_j^{-1}(\bm{x})p_t(\bm{x}),
 \end{aligned}
\end{equation}
where $C_{R,j}$ is the normalization constant, i.e. 
\begin{equation*}
C_{R,j} := \int_{\mathbb{R}^d} R_j^{-1}(\bm{x})p_t(\bm{x})\,\mathrm{d}\bm{x} =p_t\left[R_j^{-1}\right] > 0.
\end{equation*}
Meanwhile, we introduce a pseudo-gain vector field 
\begin{equation}
\label{eq:pseudo_gain_def}
\tilde{\bm{K}}_{\cdot j}(\bm{x}) \triangleq R_j(\bm{x})\bm{K}_{\cdot j}(\bm{x}) \in \mathbb{R}^d.
\end{equation}
Substituting \eqref{eq:tilde_p_def} and \eqref{eq:pseudo_gain_def} into \eqref{eq:hetero_poisson},  dividing both sides by the positive constant $C_{R,j}$ converts \eqref{eq:hetero_poisson} into a standard EL-BVP under the transformed measure $\tilde{p}_{t,j}(\bm{x})$:
\begin{equation}
\label{eq:transformed_poisson}
\nabla^T\big(\tilde{p}_{t,j}(\bm{x})\tilde{\bm{K}}_{\cdot j}(\bm{x})\big) = -\big(h_j(\bm{x}) - \tilde p_{t,j}[{h}_{j}]\big)\tilde{p}_{t,j}(\bm{x}),
 \end{equation}
where \lx{the constant $\tilde{p}_{t,j}\big[h_j\big] = \int_{\mathbb{R}^d} h_j(\bm{x})\tilde{p}_{t,j}(\bm{x})\,\mathrm{d}\bm{x}$ is determined by} the vanishing flux boundary condition \eqref{eq:std_bc_inf}. Furthermore, $\tilde p_{t,j}[{h}_{j}]$ can be represented under the original density $p_t$:
\begin{align}
\label{eq:hat_hw_continuous}
\tilde{p}_{t,j}\big[h_j\big] =& \int_{\mathbb{R}^d} h_j(\bm{x})\tilde{p}_{t,j}(\bm{x})\,\mathrm{d}\bm{x} \\\notag
=& \frac{\int_{\mathbb{R}^d} h_j(\bm{x})R_j^{-1}(\bm{x})p_t(\bm{x})\,\mathrm{d}\bm{x}}{\int_{\mathbb{R}^d} R_j^{-1}(\bm{x})p_t(\bm{x})\,\mathrm{d}\bm{x}}
=\frac{{p_t}\big[h_jR_j^{-1}\big]}{{p_t}\big[R_j^{-1}\big]}.
\end{align}

This measure transformation provides two key benefits: 1) the non-polynomial factor $R_j^{-1}(\bm{x})$ is completely absorbed into $\tilde{p}_{t,j}(\bm{x})$, restoring the polynomial form of $h_j(\bm{x})$ in \eqref{eq:transformed_poisson}; and 2) once the pseudo-gain $\tilde{\bm{K}}_{\cdot j}(\bm{x})$ is solved from \eqref{eq:transformed_poisson}, the feedback gain $\bK_{\cdot j}$ we want is algebraically recovered via \eqref{eq:pseudo_gain_def}.

\subsubsection{\lx{Importance-Reweighted GM Approximation}}

\lx{Recall the controlled particle propagation \eqref{eq:fpf_particle_sde} in the FPF, where the particles are drawn from the density $p_t(\bx)$, not the transformed density $\tilde p_t(\bx)$. To represent the transformed density with } the prior particle ensemble $\{\bm{X}_t^i\}_{i=1}^{N_p}\sim p_t(\bm{x})$ without resampling or altering the particle coordinates, we invoke importance sampling (IS). Specifically, for each $j\in\{1,\dots,m\}$ the unnormalized importance weight of the $i$-th particle $\bm{X}_t^i$ is given by the Radon-Nikodym derivative between $\tilde{p}_{t,j}(\bm{x})$ and $p_t(\bm{x})$ evaluated at $\bm{x} = \bm{X}_t^i$:
\begin{equation}
\label{eq:rn_weight_prop}
w_j^i \propto \frac{\tilde{p}_{t,j}(\bm{X}_t^i)}{p_t(\bm{X}_t^i)} = \frac{C_{R,j}^{-1}R_j^{-1}(\bm{X}_t^i)p_t(\bm{X}_t^i)}{p_t(\bm{X}_t^i)} \propto R_j^{-1}(\bm{X}_t^i),
\end{equation}
where $\propto$ means proportional to. Normalizing \eqref{eq:rn_weight_prop} across the $N_p$ particles defines the importance weight vector $\bm{w}_j = [w_j^1, \dots, w_j^{N_p}]^T \in \mathbb{R}^{N_p}$ with 
\begin{equation}
\label{eq:particle_weights}
w_j^i = \frac{R_j^{-1}(\bm{X}_t^i)}{\sum_{n=1}^{N_p} R_j^{-1}(\bm{X}_t^n)}, \quad i = 1, \dots, N_p,
\end{equation}
which satisfies $w_j^i > 0$ and $\sum_{i=1}^{N_p} w_j^i = 1$. It is worth noting that particles residing in regions with high measurement noise variance \(R_j(\bm{X}_t^i)\) are assigned small weights \(w_j^i \ll 1/N_p\), which naturally suppresses noisy outliers. For state-independent observation noise (\(R_j(\bm{x}) \equiv R_0\)), the weight vector simplifies to \(w_j^i = 1/N_p\), and \eqref{eq:weighted_gmm} reduces to the unweighted GM in \eqref{eq:unweighted_gmm}.

Using the weighted particle ensemble $\{(\bm{X}_t^i, w_j^i)\}_{i=1}^{N_p}$ and the diagonal bandwidth matrix $\bm{\Sigma} = \mathrm{diag}(\epsilon_1, \dots, \epsilon_d)$ in \eqref{eq:gaussian_kernel_def}, we construct a weighted GM approximation $\tilde{p}_{w,j}(\bm{x})$ for $\tilde{p}_{t,j}(\bm{x})$:
\begin{equation}
\label{eq:weighted_gmm}
\tilde p_{t,j}(\bx)\approx\tilde{p}_{w,j}(\bm{x}) := \sum_{i=1}^{N_p} w_j^i \mathcal{N}_i(\bm{x}),
\end{equation}
where $\N_i(\bx)$ is short for the Gaussian $\N(\bx;\bX_t^i,\bm\Sigma)$. 
Under $\tilde{p}_{w,j}(\bm{x})$, the transformed expectation $\tilde p_{w,j}[{h}_{j}]$ in \eqref{eq:hat_hw_continuous} is approximated by 
\[
    \tilde p_{w,j}[h_j] := \int_{\mathbb{R}^d} h_j(\bm{x})\tilde{p}_{w,j}(\bm{x})\,\mathrm{d}\bm{x} \approx \sum_{i=1}^{N_p} w_j^i h_j(\bm{X}_t^i),
\]
\lx{though we shall use \eqref{eq:weighted_consistency} deduced in Theorem \ref{thm:main_gain} later, instead of this approximation.}

\subsubsection{\lx{Weighted Decomposition Method}}

Expressing the $j$-th pseudo-gain column as the gradient of a scalar potential function, $\tilde{\bm{K}}_{\cdot j}(\bm{x}) = \nabla\varphi_j(\bm{x})$, and substituting the weighted GM density $\tilde{p}_{w,j}(\bm{x})$ from \eqref{eq:weighted_gmm} into \eqref{eq:transformed_poisson} yields the weighted Poisson's equation for channel $j \in \{1, \dots, m\}$:
\begin{equation}
\label{eq:weighted_poisson_gmm}
\nabla^T\big(\tilde{p}_{w,j}(\bm{x})\nabla\varphi_j(\bm{x})\big) = -\big(h_j(\bm{x}) - \tilde p_{w,j}[h_j]\big)\tilde{p}_{w,j}(\bm{x}).
\end{equation}

\lx{We now generalize the decomposition method from Section~\ref{sec-2.3} to the non-uniform importance-weighted GM approximation in \eqref{eq:weighted_poisson_gmm}, beginning with the decomposition of two sub-equations, analogous to Proposition~\ref{prop:prelim_decomp}.}
\begin{proposition}[\lx{Two sub-equations under non-uniform weighted GM}]
\label{prop:weighted_decomp}
For each observation $j \in \{1, \dots, m\}$, let the $j$-th column of the pseudo-gain be represented as $\tilde{\bm{K}}_{\cdot j}(\bm{x}) = \nabla\varphi_j(\bm{x})$. Given an ensemble of $N_p$ particles $\{\bX_t^i\}_{i=1}^{N_p}$ carrying normalized importance weights $\bm{w}_j = [w_j^1, \dots, w_j^{N_p}]^T$, the potential gradient
\begin{equation}
\label{eq:weighted_gain_reconstruct}
\begin{aligned}
\nabla\varphi_j(\bm{x}) &= \frac{1}{\sum_{i=1}^{N_p} w_j^i \mathcal{N}_i(\bm{x})} \\
&\quad \cdot \sum_{i=1}^{N_p} w_j^i \Big[\nabla^{\bm{\Sigma}}\psi_j^i(\bm{x}) + \mathcal{N}_i(\bm{x})\nabla\varphi_j^i(\bm{x})\Big],
\end{aligned}
\end{equation}
  solves the weighted Poisson's equation \eqref{eq:weighted_poisson_gmm}, where $\nabla^{\bm{\Sigma}}\psi_j^i(\bm{x})$ is the bandwidth-scaled gradient defined in \eqref{eq:inv_eig_grad_op}, and $\varphi_j^i(\bm{x})$ and $\psi_j^i(\bm{x})$ satisfy \eqref{eq:prelim_sub_pde_1} and
\begin{align}
\nabla^T\big(\nabla^{\bm{\Sigma}}\psi_j^i(\bm{x})\big) &= \big(\tilde p_{w,j}[h_j]- C_j^i\big)\mathcal{N}_i(\bm{x}), \label{eq:sub_pde_const}
\end{align}
respectively, for any constant $C_j^i \in \mathbb{R}$.
\end{proposition}
\begin{remark}
\label{rem:invariance}
Proposition \ref{prop:weighted_decomp} reveals that the non-uniform importance weights \(\{w_j^i\}_{i=1}^{N_p}\) enter only through the combination \eqref{eq:weighted_gain_reconstruct} and the term \(\tilde p_{w,j}[h_j]\) in \eqref{eq:sub_pde_const}, while the sub-equation for $\varphi_j^i(\bx)$ remains identical to \eqref{eq:prelim_sub_pde_1} from Proposition~\ref{prop:prelim_decomp} and is fully independent of \(w_j^i\).
\end{remark}
\begin{proof} \lx{The proof is similar to that of Proposition III.1, \cite{WML:25}. Given $j=1,\cdots,m$, we have
   \begin{align*}
    &\nabla^T\left(\tilde p_{w,j}(\bm{x})\nabla\varphi_j(\bx)\right)\\
    \overset{\eqref{eq:weighted_poisson_gmm}}
   =&\sum_{i=1}^{N_p}-\left(h_j(\bx)-\tilde p_{w,j}[h_j]\right)w_j^i\N_i(\bx)\\
    \overset{\eqref{eq:prelim_sub_pde_1},\eqref{eq:sub_pde_const}}=&\sum_{i=1}^{N_p}w_j^i\nabla^T\left(\mathcal{N}_i(\bx)\nabla\varphi^i_j(\bx)\right)+\sum_{i=1}^{N_p}w_j^i\nabla^T\left(\nabla^{\bm{\Sigma}}\psi_j^i(\bx)\right)\\
    =&\nabla^T\left[\sum_{i=1}^{N_p}w_j^i\left(\mathcal{N}_i(\bx)\nabla\varphi^i_j(\bx)+\nabla^{\bm{\Sigma}}\psi_j^i(\bx)\right)\right].
\end{align*}
Consequently, \eqref{eq:weighted_gain_reconstruct} follows immediately with \eqref{eq:weighted_gmm}.}
\end{proof}

By combining Proposition~\ref{prop:weighted_decomp}, the sub-equation solutions in Section~\ref{sec-2.3}, and \eqref{eq:pseudo_gain_def}, we establish the main theorem of this paper.

\begin{theorem}[Closed-form gain function in FPF-DM for the state-dependent multiplicative noise]
\label{thm:main_gain}
Consider the $d$-dimensional NLF system \eqref{eq:std_nlf_sde_a} and \eqref{eq:mult_nlf_sde} under Assumption \ref{as-1}-\ref{as-2}. Under the measure transformation \eqref{eq:pseudo_gain_def} and the weighted GM approximation \eqref{eq:weighted_gmm}, the closed-form solution for the $j$-th column of the feedback gain matrix $\bm{K}(\bm{x}) \in \mathbb{R}^{d\times m}$ satisfying the vanishing flux boundary condition \eqref{eq:std_bc_inf} is given by:
\begin{align}\label{eq:final_gain_closed_form}
\bm{K}_{\cdot j}(\bm{x}) &= R_j^{-1}(\bm{x})\frac{1}{\sum_{i=1}^{N_p} w_j^i \mathcal{N}_i(\bm{x})} \\\notag
&\quad \cdot \sum_{i=1}^{N_p} w_j^i \Bigg[ (\bm{x} - \bm{X}_t^i)\frac{\tilde p_{w,j}[h_j] - C_j^i}{2\pi^{d/2}|\bm{\Sigma}|^{1/2}}\gamma\left(\frac{d}{2}, \frac{r_i^2}{2}\right)r_i^{-d} \\\notag
&\phantom{aaaa} + \mathcal{N}_i(\bm{x})\sum_{\bm{k} \in \Omega} \tilde{\Phi}_{j,\bm{k}}^i \begin{pmatrix} 2k_1 H_{\bm{k}-\bm{e}_1}(\bm{x}) \\ \vdots \\ 2k_d H_{\bm{k}-\bm{e}_d}(\bm{x}) \end{pmatrix} \Bigg],
\end{align}
where $w_j^i$ are the importance weights \eqref{eq:particle_weights}, $r_i$ is the weighted radial distance \eqref{eq:radial_dist_r}, the coefficients $\{\tilde{\Phi}_{j,\bm{k}}^i\}_{1 \le |\bm{k}|_1 \le p}$ and constants $C_j^i$ are computed via \eqref{eq:backward_recursion}--\eqref{eq:Ci_formula}, and the constant 
\begin{equation}
\label{eq:weighted_consistency}
    \tilde p_{w,j}[h_j] = \sum_{i=1}^{N_p} w_j^i C_j^i.
\end{equation}
\end{theorem}

\begin{proof}
By Proposition~\ref{prop:weighted_decomp}, solving the weighted Poisson's equation \eqref{eq:weighted_poisson_gmm} reduces to solving the decoupled sub-equations \eqref{eq:prelim_sub_pde_1} and \eqref{eq:sub_pde_const}.

By Remark~\ref{rem:invariance}, \eqref{eq:prelim_sub_pde_1} is identical to the unweighted sub-equation \eqref{eq:prelim_sub_pde_1} in Section~\ref{sec-2.3}. Consequently, the solution $\varphi_j^i(\bm{x})$ in \eqref{eq:phi_hermite_exp} directly applies, where the coefficients $\{\tilde{\Phi}_{j,\bm{k}}^i\}_{1 \le |\bm{k}|_1 \le p}$ and the decoupling constant $C_j^i$ are given by \eqref{eq:backward_recursion} and \eqref{eq:Ci_formula}, respectively.

Second, in contrast to the unweighted sub-equation \eqref{eq:prelim_sub_pde_2}, the constant-source sub-equation \eqref{eq:sub_pde_const} involves the importance-weighted expectation $\tilde{p}_{w,j}[h_j]$. To determine $\tilde{p}_{w,j}[h_j]$, we enforce the vanishing flux boundary condition \eqref{eq:std_bc_inf} by integrating \eqref{eq:weighted_poisson_gmm} over $\mathbb{R}^d$ via the Gauss divergence theorem:
\begin{equation}
\label{eq:weighted_flux_int}
\int_{\mathbb{R}^d} \nabla^T\big(\tilde{p}_{w,j}(\bm{x})\nabla\varphi_j(\bm{x})\big)\,\mathrm{d}\bm{x} = 0.
\end{equation}
Substituting \eqref{eq:weighted_gmm} into \eqref{eq:weighted_flux_int}, the integral  becomes
\begin{equation*}
\sum_{i=1}^{N_p} w_j^i \int_{\mathbb{R}^d} \big(\tilde{p}_{w,j}[h_j] - C_j^i\big)\mathcal{N}_i(\bm{x})\,\mathrm{d}\bm{x} = 0,
\end{equation*}
which simplifies to $\sum_{i=1}^{N_p} w_j^i(\tilde{p}_{w,j}[h_j] - C_j^i) = 0$, yielding \eqref{eq:weighted_consistency}. 

Replacing $p_t^{N_p, \bm{\Sigma}}[h_j]$ with $\tilde{p}_{w,j}[h_j]$ in the radial solution \eqref{eq:psi_grad_unweighted} yields the closed-form radial gradient:
\begin{equation}
\label{eq:psi_grad_closed}
\nabla^{\bm{\Sigma}}\psi_j^i(\bm{x}) = (\bm{x} - \bm{X}_t^i)\frac{\tilde{p}_{w,j}[h_j] - C_j^i}{2\pi^{d/2}|\bm{\Sigma}|^{1/2}}\gamma\left(\frac{d}{2}, \frac{r_i^2}{2}\right)r_i^{-d}.
\end{equation}
Finally, substituting $\nabla\varphi_j^i(\bm{x})$ and \eqref{eq:psi_grad_closed} into \eqref{eq:weighted_gain_reconstruct} and applying \eqref{eq:pseudo_gain_def} yields \eqref{eq:final_gain_closed_form}.
\end{proof}

The complete numerical implementation of the proposed multivariate FPF-DM for multiplicative observation noise is summarized in Algorithm~\ref{alg:proposed_mult_fpf}.

\begin{algorithm}[!t]
\caption{Multivariate FPF-DM for Multiplicative Observation Noise}
\label{alg:proposed_mult_fpf}
\begin{algorithmic}[1]
\REQUIRE Initial prior density $p_0^*(\bm{x})$, particle count $N_p$, bandwidth matrix $\bm{\Sigma} = \mathrm{diag}(\epsilon_1, \dots, \epsilon_d)$, time step $\Delta t$, total steps $N_t = T/\Delta t$, Hermite coefficients $\{a_{j,\bm{k}}\}$, state-dependent noise variance functions $\{R_j(\bm{x})\}_{j=1}^m$.
\ENSURE Posterior state estimates $\{\hat{\bm{X}}_{t_n}\}_{n=0}^{N_t}$.
\STATE \textbf{Initialization:} Sample initial particles $\{\bm{X}_0^i\}_{i=1}^{N_p} \stackrel{\mathrm{i.i.d.}}{\sim} p_0^*(\bm{x})$.
\FOR{$n = 0$ \textbf{to} $N_t - 1$}
    \FOR{each observation channel $j = 1$ \textbf{to} $m$}
        \STATE Evaluate $R_j(\bm{X}_{t_n}^i)$ and compute the normalized importance weight vector $\bm{w}_j = [w_j^1, \dots, w_j^{N_p}]^T$ via \eqref{eq:particle_weights}
        \FOR{each particle $i = 1$ \textbf{to} $N_p$}
            \STATE Set $\tilde{\Phi}_{j,\bm{k}}^i = 0$ for $|\bm{k}|_1 > p$, and compute $\tilde{\Phi}_{j,\bm{k}}^i$ for $|\bm{k}|_1 = p, \dots, 1$ via backward recursion \eqref{eq:backward_recursion}.
            \STATE Compute the decoupling constant $C_j^i$ via \eqref{eq:Ci_formula}.
        \ENDFOR
        \STATE Compute the precision-weighted global innovation $\tilde{p}_{w,j}[h_j] = \sum_{i=1}^{N_p} w_j^i C_j^i$.
        \STATE Evaluate the $j$-th gain column $\bm{K}_{\cdot j}(\bm{X}_{t_n}^i)$ at each particle location $\bm{x} = \bm{X}_{t_n}^i$ via \eqref{eq:final_gain_closed_form}.
    \ENDFOR
    \FOR{each particle $i = 1$ \textbf{to} $N_p$}
        \STATE Generate Wiener process increment $\Delta\bm{B}_{t_n}^i \sim \mathcal{N}(\bm{0}, \Delta t \bm{I}_d)$.
        \STATE Compute particle innovation $\Delta\bm{I}_{t_n}^i = \Delta\bm{Z}_{t_n} - \frac{1}{2}\big(\bm{h}(\bm{X}_{t_n}^i) + \tilde{\bm{p}}_w[\bm{h}]\big)\Delta t$, where $\tilde{\bm{p}}_w[\bm{h}] = \big[\tilde{p}_{w,1}[h_1], \dots, \tilde{p}_{w,m}[h_m]\big]^T$.
        \STATE Propagate and update the particle state via Euler discretization:
        \[
        \begin{aligned}
            \bm{X}_{t_{n+1}}^i &= \bm{X}_{t_n}^i + \bm{g}(\bm{X}_{t_n}^i)\Delta t + \bm{\sigma}(\bm{X}_{t_n}^i)\Delta\bm{B}_{t_n}^i \\
            &\quad + \bm{K}(\bm{X}_{t_n}^i)\Delta\bm{I}_{t_n}^i.
        \end{aligned}
        \]
    \ENDFOR
    \STATE Compute posterior state estimate $\hat{\bm{X}}_{t_{n+1}} = \frac{1}{N_p}\sum_{i=1}^{N_p} \bm{X}_{t_{n+1}}^i$.
\ENDFOR
\end{algorithmic}
\end{algorithm}

\subsection{\lx{The QFPF-DM and DQFPF-DM}}
\label{sec-3.3}

We now specialize the FPF-DM framework developed in Section~\ref{sec:measure_trans} (Theorem~\ref{thm:main_gain} and Algorithm~\ref{alg:proposed_mult_fpf}) to the attitude and pose estimation problems formulated in Section~\ref{sec-2.1}. In both systems \eqref{eq:4d_sde_a} and \eqref{eq:8d_sde_a}, the rotational state evolves on $\mathcal{S}^3 \subset \mathbb{R}^4$ ($d = 4$) with $m = 4$ observations  and an isotropic bandwidth matrix $\bm{\Sigma} = \epsilon\bm{I}_4$ ($\epsilon > 0$). Substituting $d = 4$ and $\bm{\Sigma} = \epsilon\bm{I}_4$ into \eqref{eq:final_gain_closed_form}, we have $|\bm{\Sigma}|^{1/2} = \epsilon^2$, $r_i(\tilde{\bm{q}}) = \|\tilde{\bm{q}} - \tilde{\bm{q}}_t^i\|_2 / \sqrt{\epsilon}$, and the lower incomplete Gamma function $\gamma(2, r_i^2/2) = 1 - (1 + r_i^2/2)\exp(-r_i^2/2)$. 

\yjy{Furthermore, unlike the unconstrained Euclidean update in Algorithm~\ref{alg:proposed_mult_fpf}, the unit quaternion states in \eqref{eq:4d_sde_a} and \eqref{eq:8d_sde_a} must satisfy the spherical manifold constraint $\mathcal{S}^3 = \{\tilde{\bm{q}} \in \mathbb{R}^4 : \|\tilde{\bm{q}}\|_2 = 1\}$. Therefore, after the Euler discretization update (Line~15 of Algorithm~\ref{alg:proposed_mult_fpf}), each quaternion particle and the weighted posterior estimate are normalized back onto $\mathcal{S}^3$ via $\tilde{\bm{q}}_{t_{n+1}}^i \leftarrow \tilde{\bm{q}}_{t_{n+1}}^i / \|\tilde{\bm{q}}_{t_{n+1}}^i\|_2$ and $\tilde{\bm{q}}_{\mathrm{est},t_{n+1}} \leftarrow \tilde{\bm{q}}_{\mathrm{est},t_{n+1}} / \|\tilde{\bm{q}}_{\mathrm{est},t_{n+1}}\|_2$ \cite{li2022pose}.}

In the following two subsections, we specialize the Hermite backward recursion \eqref{eq:backward_recursion}--\eqref{eq:Ci_formula} in Theorem~\ref{thm:main_gain} to $p = 2$ and $p = 1$, yielding the QFPF-DM and DQFPF-DM algorithms:

\subsubsection{QFPF-DM for 4D Attitude Tracking ($p = 2$):} 
\yjy{For the quadratic observation $h_j(\tilde{\bm{q}}) = \tilde{\bm{q}}^T\bm{S}_j\tilde{\bm{q}}$ ($j = 1, \dots, 4$) in \eqref{eq:4d_sde_b}, expanding $h_j(\tilde{\bm{q}})$ in the 4D Hermite polynomials \eqref{eq:hermite_obs_assumption} ($H_0(q_q) = 1$, $H_1(q_q) = 2q_q$, $H_2(q_q) = 4q_q^2 - 2$) gives the non-zero coefficients $a_{j,2\bm{e}_l} = \frac{1}{4}S_{j,ll}$, $a_{j,\bm{e}_l+\bm{e}_q} = \frac{1}{2}S_{j,lq}$ ($l < q$), and $a_{j,\bm{0}} = \frac{1}{2}\mathrm{tr}(\bm{S}_j)$, where $S_{j,lq}$ is the $(l, q)$-th entry of $\bm{S}_j$. Substituting these into \eqref{eq:backward_recursion}--\eqref{eq:Ci_formula} yields the closed-form parameters for each particle $\tilde{\bm{q}}_t^i$:
    \begin{equation}
    \label{eq:qfpf_hermite_coeffs}
    \begin{aligned}
    \tilde{\Phi}_{j,2\bm{e}_l}^i &= \frac{\epsilon}{8}S_{j,ll}, \\
    \tilde{\Phi}_{j,\bm{e}_l}^i &= \frac{\epsilon}{2}\sum_{q=1}^{4} S_{j,lq}q_{t,q}^i, \\
    \tilde{\Phi}_{j,\bm{e}_l+\bm{e}_q}^i &= \frac{\epsilon}{4}S_{j,lq} \quad (l < q), \\
    C_j^i &= (\tilde{\bm{q}}_t^i)^T\bm{S}_j\tilde{\bm{q}}_t^i + \epsilon\,\mathrm{tr}(\bm{S}_j),
    \end{aligned}
    \end{equation}
    and the local polynomial potential gradient in \eqref{eq:final_gain_closed_form} simplifies to the linear vector field 
    \begin{equation}\label{eqn-3.1}
    \nabla\varphi_j^i(\tilde{\bm{q}}) = \epsilon\bm{S}_j(\tilde{\bm{q}} + \tilde{\bm{q}}_t^i).
    \end{equation}
    Evaluating the feedback gain $\bm{K}(\tilde{\bm{q}}_t^i)$ via \eqref{eq:final_gain_closed_form} with \eqref{eq:qfpf_hermite_coeffs} and propagating the quaternion particles on $\mathcal{S}^3$ via Algorithm~\ref{alg:proposed_mult_fpf} establishes the QFPF-DM algorithm.}

\subsubsection{DQFPF-DM for 8D Pose Estimation ($p = 1$):} 

\yjy{For the pseudo-linear observation $h_{k,j}(\tilde{\bm{q}}_r) = \sum_{q=1}^{4} M_{k,jq} q_q$ ($j = 1, \dots, 4$) in \eqref{eq:8d_sde_b}, where $M_{k,jq}$ is the $(j, q)$-th entry of $\bm{M}_k$ in \eqref{eq:Mk_def}, the non-zero Hermite expansion coefficients in \eqref{eq:hermite_obs_assumption} are $a_{j,\bm{e}_q} = \frac{1}{2}M_{k,jq}$ ($q = 1, \dots, 4$) and $a_{j,\bm{0}} = 0$. Since all higher-order coefficients $|\bm{k}|_1 \ge 2$ vanish for $p = 1$, the backward recursion \eqref{eq:backward_recursion}--\eqref{eq:Ci_formula} reduces to:
\begin{equation}\label{eq:dqfpf_hermite_coeffs}
\begin{aligned}
    \tilde{\Phi}_{j,\bm{e}_q}^i &= \frac{\epsilon}{2}M_{k,jq},\\
    \quad C_j^i &= \sum_{q=1}^{4} M_{k,jq} q_{r,t,q}^i = h_{k,j}(\tilde{\bm{q}}_{r,t}^i),
    \end{aligned}
\end{equation}
    and the local polynomial potential gradient in \eqref{eq:final_gain_closed_form} reduces to the constant vector $\nabla\varphi_j^i(\tilde{\bm{q}}_r) = \epsilon\bm{M}_{k,j\cdot}^T$, where $\bm{M}_{k,j\cdot}$ is the $j$-th row of $\bm{M}_k$. Evaluating the gain via \eqref{eq:final_gain_closed_form} with \eqref{eq:dqfpf_hermite_coeffs} and executing Algorithm~\ref{alg:proposed_mult_fpf} updates the rotation estimate $\tilde{\bm{q}}_{r,\mathrm{est},k}$ on $\mathcal{S}^3$, after which the translation $\bm{t}_{\mathrm{est},k}$ and dual part $\tilde{\bm{q}}_{d,\mathrm{est},k}$ are recovered via \eqref{eq:dq_sum_trans_eq}, establishing the DQFPF-DM algorithm.}

\section{NUMERICAL SIMULATIONS}\label{sec-4}

In this section, numerical simulations are conducted on the two estimation problems formulated in Section~\ref{sec-2.1} to evaluate the estimation accuracy and computational efficiency of the proposed QFPF-DM and DQFPF-DM algorithms established in Section~\ref{sec-3.3}: 1) the 4D gyroscope-driven quaternion attitude tracking problem ($p = 2$); and 2) the 8D dual-quaternion point-cloud registration problem ($p = 1$).

In both experiments, QFPF-DM and DQFPF-DM are benchmarked against three baselines: constant-gain FPF (QFPF-CG in 4D, DQFPF-CG in 8D) \cite{li2022pose}; sequential Monte Carlo PF (QPF in 4D, DQPF in 8D); and Kalman-filter variants (QEKF in 4D \cite{choukroun2006novel}, linear DQKF in 8D \cite{srivatsan2016estimating}). To demonstrate that the proposed method achieves superior accuracy with a significantly smaller ensemble, the particle counts in both experiments are set to $N_p = 10$ for the QFPF-DM and DQFPF-DM, $N_p = 50$ for the QFPF-CG and DQFPF-CG, and $N_p = 200$ for the QPF and DQPF. All simulations are executed in MATLAB R2022a, and 100 independent Monte Carlo (MC) trials are performed for each scenario.

\subsection{4D Quaternion Gyroscope Attitude Tracking}
\label{sec:sim_4d}

For the 4D quaternion attitude tracking system \eqref{eq:4d_sde_a} and \eqref{eq:4d_sde_b} in Section~\ref{sec-2.1}, we consider $m = 4$ quadratic observations ($p = 2$) with the symmetric observation matrices $\bm{S}_1, \dots, \bm{S}_4 \in \mathbb{R}^{4\times 4}$ specified as:
\begin{equation}
\label{eq:S_matrices_4d}
\begin{aligned}
\bm{S}_1 &= \mathrm{diag}(1, \, 1, \, -1, \, -1), \quad \bm{S}_2 = \mathrm{diag}(1, \, -1, \, 1, \, -1), \\
\bm{S}_3 &= \begin{bmatrix} 0 & 0 & 1 & 0 \\ 0 & 0 & 0 & 1 \\ 1 & 0 & 0 & 0 \\ 0 & 1 & 0 & 0 \end{bmatrix}, \quad \bm{S}_4 = \begin{bmatrix} 0 & 1 & 0 & 0 \\ 1 & 0 & 0 & 0 \\ 0 & 0 & 0 & -1 \\ 0 & 0 & -1 & 0 \end{bmatrix}.
\end{aligned}
\end{equation}
\yjy{In \eqref{eq:4d_sde_a} and \eqref{eq:4d_sde_b}, the gyroscope noise standard deviation is set to $\sigma_g = 0.008\text{ rad}/\sqrt{\text{s}}$, and the isotropic kernel bandwidth of QFPF-DM is set to $\bm{\Sigma} = \epsilon\bm{I}_4$ ($\epsilon = 1.5 \times 10^{-4}$). Since $\|\bm{q}_{v,t}\|_2^2 = \sin^2(\theta_t/2)$ quantifies the attitude inclination angle from the reference frame, the state-dependent observation noise variance in (4) is modeled as
\begin{equation*}
R_j(\tilde{\bm{q}}_t, t) = R(\tilde{\bm{q}}_t) = R_0\big(1 + \alpha\|\bm{q}_{v,t}\|_2^2\big),
\end{equation*}
with nominal variance $R_0 = 4 \times 10^{-5}$ and state-sensitivity coefficient $\alpha = 45$.} 

\yjy{In the simulation, the true attitude is initialized at $\tilde{\bm{q}}_{\mathrm{true},0} = [1, \, 0, \, 0, \, 0]^T$ and driven by $\bm{\omega}_{\mathrm{true},t} = [0.08\sin(0.06t), \, 0.07\cos(0.05t), \, 0.05\sin(0.04t)]^T\text{ rad/s}$ for $T = 100\text{ s}$ with step size $\Delta t = 0.01\text{ s}$ ($N_t = 10000$ steps).} All estimators are initialized at the perturbed state $\tilde{\bm{q}}_{\mathrm{est},0} = [\cos(0.03), \, \sin(0.03), \, -0.015, \, 0.01]^T / \|\cdot\|_2$, corresponding to an initial attitude error of approximately $4^\circ$, with initial particles sampled around $\tilde{\bm{q}}_{\mathrm{est},0}$ with a standard deviation of $0.008$. At each discrete time step $k$, the geodesic attitude angular error $\Delta\theta_k$ (in degrees) on $SO(3)$ is defined by:
\begin{equation}
\label{eq:geodesic_err_def}
\Delta\theta_k = \frac{360^\circ}{\pi}\arccos\Big(\big|\tilde{\bm{q}}_{\mathrm{est},k}^T\tilde{\bm{q}}_{\mathrm{true},k}\big|\Big).
\end{equation}
Over the steady-state tracking window $k \in \{k_0, \dots, N_t\}$ with $k_0 = 40$ and $N_{\mathrm{ss}} = N_t - k_0 + 1$, filter performance is evaluated using the attitude root mean square error ($\mathrm{RMSE}_\theta$) and the Euclidean quaternion root mean square error ($\mathrm{RMSE}_{\tilde{\bm{q}}}$), defined as:
\begin{align}
\mathrm{RMSE}_\theta &= \sqrt{\frac{1}{N_{\mathrm{ss}}}\sum_{k=k_0}^{N_t} (\Delta\theta_k)^2}, \label{eq:att_rmse_def} \\
\mathrm{RMSE}_{\tilde{\bm{q}}} &= \sqrt{\frac{1}{N_{\mathrm{ss}}}\sum_{k=k_0}^{N_t} \big\|\tilde{\bm{q}}_{\mathrm{est},k} - \tilde{\bm{q}}_{\mathrm{true},k}\big\|_2^2}, \label{eq:quat_rmse_def}
\end{align}
respectively, where the antipodal sign of $\tilde{\bm{q}}_{\mathrm{est},k}$ in \eqref{eq:quat_rmse_def} is aligned with $\tilde{\bm{q}}_{\mathrm{true},k}$ such that $\tilde{\bm{q}}_{\mathrm{est},k}^T\tilde{\bm{q}}_{\mathrm{true},k} \ge 0$.

\begin{figure}[!t]
\centerline{\includegraphics[width=0.96\columnwidth]{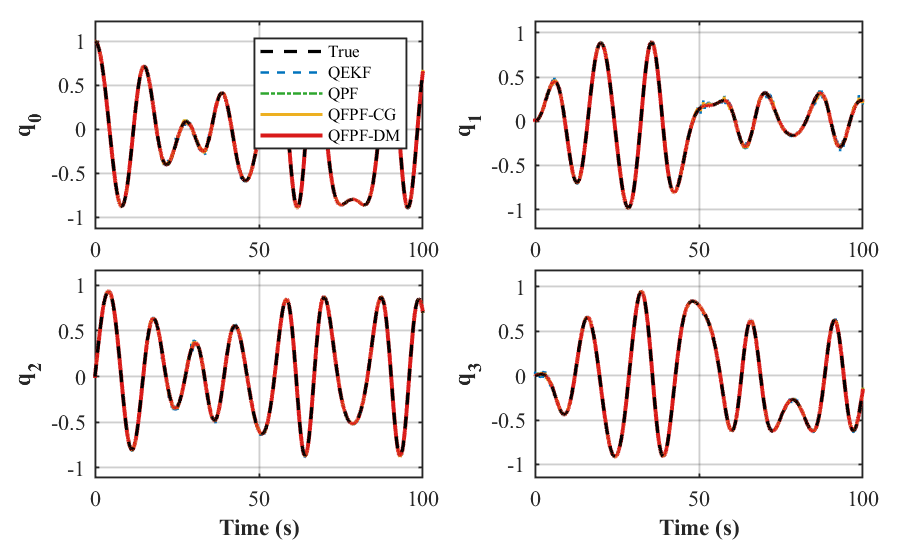}}
\caption{Tracking trajectories of the four unit quaternion components $q_0, q_1, q_2, q_3$ over time for one single run}
\label{fig:4d_tracking}
\end{figure}

\begin{figure}[!t]
\centerline{\includegraphics[width=0.95\columnwidth]{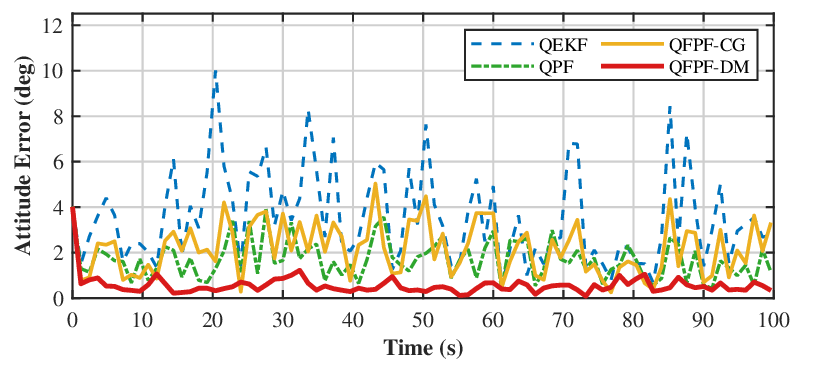}}
\caption{Geodesic attitude estimation error in $SO(3)$ versus time of a single run}
\label{fig:4d_single_err}
\end{figure}

\begin{figure}[!t]
\centerline{\includegraphics[width=0.95\columnwidth]{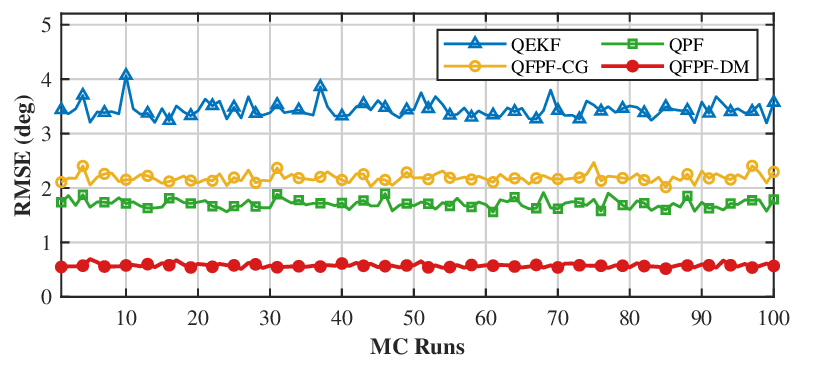}}
\caption{Comparison of steady-state geodesic attitude RMSE over 100 MC runs}
\label{fig:4d_mc_rmse}
\end{figure}

\begin{table}[!t]
\caption{4D quaternion attitude tracking: averaged RMSE errors and runtime over 100 MC runs ($T=100$ s, $N_t=10000$)}
\label{tab:4d_mc_table}
\tablefont
\centering
\setlength{\tabcolsep}{4pt}
\begin{tabular}{lcccc}
\hline
Method & $N_p$ & Att. RMSE (${}^\circ$) & Quat. RMSE & Avg Time (s) \\
\hline
QFPF-DM & $10$ & $\mathbf{0.5740}$ & $\mathbf{0.00501}$ & $0.5819$ \\
QFPF-CG & $50$ & $2.1875$ & $0.01909$ & $\underline{0.3243}$ \\
QPF     & $200$ & $\underline{1.7093}$ & $\underline{0.01492}$ & $2.6796$ \\
QEKF    & --- & $3.4369$ & $0.02999$ & $\mathbf{0.0983}$ \\
\hline
\end{tabular}
\begin{tablenotes}
\item \textit{Note:} Bold values denote the best result, and underlined values denote the second-best result for each metric.
\end{tablenotes}
\end{table}

Fig.~\ref{fig:4d_tracking} illustrates the tracking trajectories of the four quaternion components in one single run, while Fig.~\ref{fig:4d_single_err} and Fig.~\ref{fig:4d_mc_rmse} depict the single-run geodesic attitude error $\Delta\theta_k$ and the steady-state $\mathrm{RMSE}_\theta$ across 100 MC runs, respectively. The averaged estimation errors and execution times are summarized in Table~\ref{tab:4d_mc_table}. \lx{As shown in the table, the proposed QFPF-DM with only $N_p=10$ particles achieves superior estimation accuracy and smooth tracking, surpassing QFPF-CG, QPF, and QEKF. For this quadratic observation example, QFPF-CG performs worse than the 200-particle QPF since the true gain varies linearly across the state space as deduced in \eqref{eqn-3.1} of Section \ref{sec-3.3}, and the constant-gain approximation in QFPF-CG yields considerable truncation errors during attitude maneuvers, whereas QFPF-DM computes the spatially varying gain analytically. Despite a modest increase in runtime relative to QFPF-CG from Hermite backward recursion, QFPF-DM is considerably faster than the 200-particle QPF while achieving much lower estimation error.}

\subsection{8D Dual Quaternion Point Cloud Registration}
\label{sec:sim_8d}

We next evaluate the 8D dual-quaternion pose registration system \eqref{eq:8d_sde_a} and \eqref{eq:8d_sde_b}--\eqref{eq:dq_sum_trans_eq} described in Section~\ref{sec-2.1}. Following \cite{li2022pose}, $N_{\mathrm{pts}} = 400$ points are randomly sampled from the Stanford Bunny dataset. The true pose is defined by a rotation of $\theta = \pi/3\text{ rad}$ about the $x$-axis, corresponding to $\tilde{\bm{q}}_{r,\mathrm{true}} = [0.866, \, 0.500, \, 0.000, \, 0.000]^T$, and a translation of $\bm{t}_{\mathrm{true}} = [50, \, 42, \, 20]^T\text{ mm}$. All estimators are initialized at $\tilde{\bm{q}}_{r,0} = [0.866, \, 0.452, \, 0.151, \, 0.150]^T$ and $\bm{t}_0 = [0, \, 0, \, 0]^T\text{ mm}$. \yjy{At each sequential measurement step $k$, the filtering system \eqref{eq:8d_sde_a} and \eqref{eq:8d_sde_b} evolves over a pseudo-time horizon of $T = 1.0\text{ s}$ with step size $\Delta t = 0.04\text{ s}$,} and the kernel bandwidth of DQFPF-DM is set to $\bm{\Sigma} = 10^{-4}\bm{I}_4$. To reflect sensor degradation near point-cloud boundaries and state-multiplicative error coupling in \eqref{eq:8d_sde_b}, each measured point $\bm{a}_k$ is corrupted by zero-mean Gaussian noise with spatially varying variance $\sigma_k^2$ scaled by a base variance of $\sigma_0^2 = 4.0\text{ mm}^2$ in the single-run evaluation and $\sigma_0^2 \sim \mathcal{U}(1.5, \, 6.0)\text{ mm}^2$ across the 100 MC trials, \yjy{and the state-dependent observation noise variance in Section~\ref{sec-2.1} is modeled as $R_k(\tilde{\bm{q}}_{r,t}) = \bar{\sigma}_k^2 + \eta\|\bm{M}_k\tilde{\bm{q}}_{r,t}\|_2^2$ with pair variance $\bar{\sigma}_k^2 = \frac{1}{2}(\sigma_k^2 + \sigma_{k-1}^2)$ and coupling coefficient $\eta = 0.01$.} Registration accuracy at step $k$ is quantified by the point-to-point RMS:
\begin{equation}
\label{eq:8d_rms_def}
\begin{aligned}
&\mathrm{RMS}_k \\= 
&\sqrt{\frac{1}{N_{\mathrm{pts}}}\sum_{m=1}^{N_{\mathrm{pts}}} \big\|\bm{C}(\tilde{\bm{q}}_{r,\mathrm{est},k})\bm{b}_m + \bm{t}_{\mathrm{est},k} - \bm{a}_{\mathrm{true},m}\big\|_2^2},
\end{aligned}
\end{equation}
where $\bm{a}_{\mathrm{true},m} = \bm{C}(\tilde{\bm{q}}_{r,\mathrm{true}})\bm{b}_m + \bm{t}_{\mathrm{true}}$ denotes the noise-free target point.
\begin{table*}[!t]
\caption{Estimated pose parameters, final registration RMS error, and execution time on the noisy Stanford bunny dataset}
\label{tab:8d_single_table}
\tablefont
\centering
\setlength{\tabcolsep}{4pt}
\begin{tabular*}{\textwidth}{@{\extracolsep{\fill}}lcccccccccc@{}}
\hline
Method & $N_p$ & $q_0$ & $q_1$ & $q_2$ & $q_3$ & $t_x$ (mm) & $t_y$ (mm) & $t_z$ (mm) & RMS (mm) & Time (s) \\
\hline
Actual Pose & --- & $0.866$ & $0.500$ & $0.000$ & $0.000$ & $50.000$ & $42.000$ & $20.000$ & --- & --- \\
DQFPF-DM & $10$ & $0.867$ & $0.499$ & $-0.000$ & $-0.000$ & $49.989$ & $42.065$ & $20.097$ & $\mathbf{0.1732}$ & $0.4015$ \\
DQFPF-CG & $50$ & $0.866$ & $0.500$ & $-0.003$ & $0.000$ & $49.866$ & $42.010$ & $20.236$ & $0.3751$ & $\underline{0.1788}$ \\
DQPF & $200$ & $0.866$ & $0.500$ & $-0.000$ & $-0.000$ & $49.815$ & $41.991$ & $20.252$ & $\underline{0.3144}$ & $0.9991$ \\
DQKF & --- & $0.870$ & $0.493$ & $-0.003$ & $-0.006$ & $58.787$ & $46.473$ & $20.685$ & $9.9548$ & $\mathbf{0.0031}$ \\
\hline
\end{tabular*}
\begin{tablenotes}
\item \textit{Note:} Bold values denote the best result, and underlined values denote the second-best result for each metric.
\end{tablenotes}
\end{table*}

\begin{figure}[!t]
\centerline{\includegraphics[width=0.95\columnwidth]{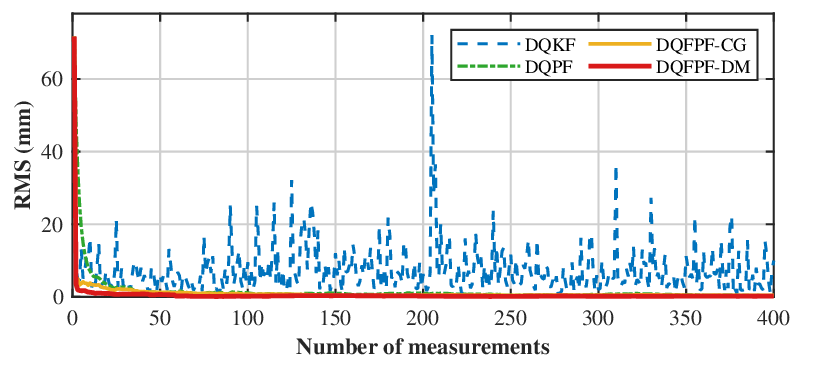}}
\caption{Point cloud registration RMS error versus the number of sequential measurements on the noisy Stanford Bunny dataset}
\label{fig:8d_single}
\end{figure}

\begin{figure}[!t]
\centerline{\includegraphics[width=0.95\columnwidth]{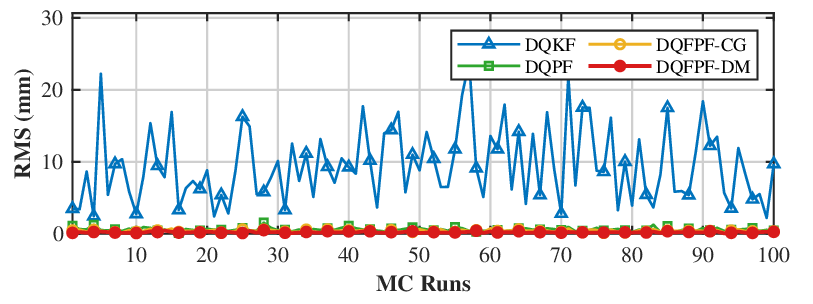}}
\vspace{1pt}
\centerline{\footnotesize (a) All Methods}
\vspace{3pt}
\centerline{\includegraphics[width=0.95\columnwidth]{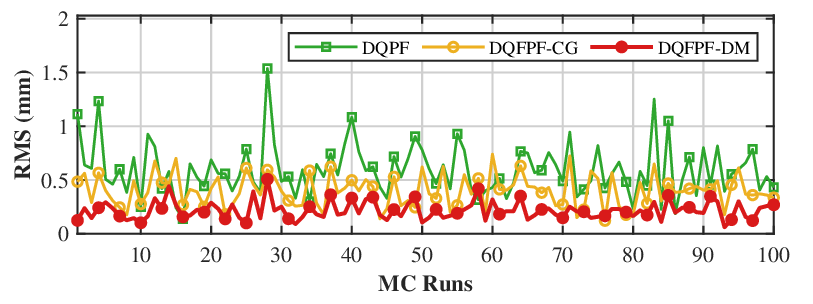}}
\vspace{1pt}
\centerline{\footnotesize (b) Zoomed-in View}
\caption{Point cloud registration RMS errors over 100 MC runs: (a) all four methods and (b) zoomed-in view of three comparative methods: DQFPF-DM, DQFPF-CG, and DQPF}
\label{fig:8d_mc}
\end{figure}

\begin{table}[!t]
\caption{Averaged performance comparison for 8D point cloud registration over 100 Monte Carlo runs}
\label{tab:8d_mc_table}
\tablefont
\centering
\setlength{\tabcolsep}{5pt}
\begin{tabular*}{\columnwidth}{@{\extracolsep{\fill}}lccc@{}}
\hline
Method & $N_p$ & Avg RMS (mm) & Avg Time (s) \\
\hline
DQFPF-DM & $10$ & $\mathbf{0.2169}$ & $0.3342$ \\
DQFPF-CG & $50$ & $\underline{0.3999}$ & $\underline{0.1809}$ \\
DQPF & $200$ & $0.5940$ & $0.9991$ \\
DQKF & --- & $9.5210$ & $\mathbf{0.0036}$ \\
\hline
\end{tabular*}
\begin{tablenotes}
\item \textit{Note:} Bold values denote the best result, and underlined values denote the second-best result for each metric.
\end{tablenotes}
\end{table}

The single-run registration results under spatial multiplicative noise are presented in Table~\ref{tab:8d_single_table} and Fig.~\ref{fig:8d_single}, while the 100 MC statistics are summarized in Table~\ref{tab:8d_mc_table} and Fig.~\ref{fig:8d_mc}. Although the linear DQKF achieves the shortest execution time, the uncompensated state-multiplicative noise induces severe estimation bias, resulting in a large mean RMS error.

Comparing Table~\ref{tab:8d_mc_table} with Table~\ref{tab:4d_mc_table} reveals a contrasting performance pattern: unlike QFPF-CG, which is outperformed by QPF in the 4D quadratic attitude estimation, DQFPF-CG yields better accuracy than DQPF for the 8D pseudo-linear registration task. This discrepancy arises because the observation mapping \(\bm{M}_k\tilde{\bm{q}}_r\) in \eqref{eq:8d_sde_b} is pseudo-linear (\(p=1\)). As derived in Section~\ref{sec-3.3}, the corresponding local potential gradient \(\nabla\varphi_j^i(\tilde{\bm{q}}_r)=\epsilon\bm{M}_{k,j\cdot}^T\) is spatially constant, so DQFPF-CG incurs no truncation error, whereas DQPF suffers from sample impoverishment. Nevertheless, DQFPF-CG remains susceptible to noisy boundary measurements. With precision-aware importance reweighting and radial Gamma correction, the proposed DQFPF-DM attains the best registration accuracy. In addition, DQFPF-DM substantially reduces computational overhead relative to DQPF while preserving the leading estimation performance.

\section{CONCLUSION}
\label{sec-5}
This paper develops a general multivariate FPF decomposition method (FPF-DM) for nonlinear systems with state-dependent multiplicative measurement noise. By means of measure transformation and importance reweighting, the state-dependent inverse noise covariance in the Poisson equation is incorporated into a non-uniformly weighted Gaussian mixture, restoring the polynomial structure and delivering closed-form feedback gains using Hermite backward recursion and incomplete Gamma functions. QFPF-DM and DQFPF-DM represent specific realizations of this broader framework for attitude and pose estimation on the unit quaternion and dual-quaternion manifolds, respectively. Simulations on 4D attitude tracking and 8D point-cloud registration verify that these realizations, using only 10 particles, outperform the QFPF-CG/DQFPF-CG, QPF/DQPF, and QEKF/DQKF baselines in estimation accuracy and are substantially faster than the QPF/DQPF.

\appendix
\section*{Preliminaries on Quaternions and Dual Quaternions}
\label{app}
To parameterize the 3D rotation group $SO(3)$ and the 6-DoF special Euclidean group $SE(3)$ without gimbal lock singularities or trigonometric nonlinearities, unit quaternions and unit dual quaternions are employed by embedding them into $\mathbb{R}^4$ \cite{choukroun2023quaternion} and $\mathbb{R}^8$ \cite{ zivan2022dual}, respectively. 

\subsubsection{Unit Quaternions}
A quaternion $\tilde{\bm{q}} \in \mathbb{R}^4$ is defined as $\tilde{\bm{q}} = [q_0, \, \bm{q}_v^T]^T = [q_0, \, q_1, \, q_2, \, q_3]^T$, where $q_0 \in \mathbb{R}$ is the scalar part and $\bm{q}_v = [q_1, \, q_2, \, q_3]^T \in \mathbb{R}^3$ is the vector part \cite{choukroun2023quaternion}. Any 3D vector $\bm{a} \in \mathbb{R}^3$ is embedded as a pure quaternion with zero scalar part, $\tilde{\bm{a}} = [0, \, \bm{a}^T]^T$. Given two quaternions $\tilde{\bm{p}} = [p_0, \, \bm{p}_v^T]^T$ and $\tilde{\bm{q}} = [q_0, \, \bm{q}_v^T]^T$, their bilinear multiplication $\odot$ can be written in matrix-vector form as \cite{li2022pose}:
\begin{equation}
\label{eq:quat_mult_def}
\tilde{\bm{p}} \odot \tilde{\bm{q}} = \bm{L}(\tilde{\bm{p}})\tilde{\bm{q}} = \bm{R}(\tilde{\bm{q}})\tilde{\bm{p}},
\end{equation}
where the $4\times 4$ left- and right-multiplication matrices $\bm{L}(\tilde{\bm{p}})$ and $\bm{R}(\tilde{\bm{q}})$ are defined by:
\begin{align}
\bm{L}(\tilde{\bm{p}}) &:= \begin{bmatrix} p_0 & -\bm{p}_v^T \\ \bm{p}_v & p_0\bm{I}_3 + [\bm{p}_v]^\times \end{bmatrix}, \label{eq:quat_L_mat} \\
\bm{R}(\tilde{\bm{q}}) &:= \begin{bmatrix} q_0 & -\bm{q}_v^T \\ \bm{q}_v & q_0\bm{I}_3 - [\bm{q}_v]^\times \end{bmatrix}, \label{eq:quat_R_mat}
\end{align}
with $\bm{I}_3$ denoting the $3\times 3$ identity matrix and $[\bm{v}]^\times$ denoting the $3\times 3$ skew-symmetric cross-product matrix for $\bm{v} = [v_1, \, v_2, \, v_3]^T$:
\begin{equation}
\label{eq:skew_sym_mat}
[\bm{v}]^\times := \begin{bmatrix} 0 & -v_3 & v_2 \\ v_3 & 0 & -v_1 \\ -v_2 & v_1 & 0 \end{bmatrix}.
\end{equation}
When $\tilde{\bm{q}}$ multiplies a pure quaternion $\tilde{\bm{v}} = [0, \, \bm{v}^T]^T$, \eqref{eq:quat_mult_def} yields the reordering identity:
\begin{equation}
\label{eq:quat_vec_reorder}
\tilde{\bm{q}} \odot \tilde{\bm{v}} = \bm{R}(\tilde{\bm{v}})\tilde{\bm{q}} = \bm{\Xi}(\tilde{\bm{q}})\bm{v},
\end{equation}
where the linear matrix function $\bm{\Xi}(\tilde{\bm{q}}) \in \mathbb{R}^{4\times 3}$ is given by:
\begin{equation}
\label{eq:Xi_matrix_def}
\bm{\Xi}(\tilde{\bm{q}}) := \begin{bmatrix} -\bm{q}_v^T \\ q_0\bm{I}_3 + [\bm{q}_v]^\times \end{bmatrix}.
\end{equation}

The conjugate of $\tilde{\bm{q}}$ is $\tilde{\bm{q}}^* = [q_0, \, -\bm{q}_v^T]^T$, satisfying $(\tilde{\bm{p}} \odot \tilde{\bm{q}})^* = \tilde{\bm{q}}^* \odot \tilde{\bm{p}}^*$ and $\tilde{\bm{a}}^* = -\tilde{\bm{a}}$ for any pure quaternion $\tilde{\bm{a}}$. A unit quaternion lies on the three-sphere $\mathcal{S}^3 := \{\tilde{\bm{q}} \in \mathbb{R}^4 : \|\tilde{\bm{q}}\|_2 = 1\}$, for which $\tilde{\bm{q}}^{-1} = \tilde{\bm{q}}^*$ and $\bm{\Xi}^T(\tilde{\bm{q}})\bm{\Xi}(\tilde{\bm{q}}) = \bm{I}_3$. Any 3D rotation by an angle $\theta \in [-\pi, \pi]$ about a unit axis $\bm{k} \in \mathbb{R}^3$ is parameterized by $\tilde{\bm{q}} = [\cos(\theta/2), \, \sin(\theta/2)\bm{k}^T]^T$, where $\tilde{\bm{q}}$ and $-\tilde{\bm{q}}$ represent the same attitude.

\subsubsection{Unit Dual Quaternions}
A dual quaternion $\hat{\bm{q}} \in \mathbb{R}^8$ is defined as \cite{zivan2022dual, srivatsan2016estimating}:
\begin{equation}
\label{eq:dq_def}
\hat{\bm{q}} = \tilde{\bm{q}}_r + \varepsilon\tilde{\bm{q}}_d = [\tilde{\bm{q}}_r^T, \, \tilde{\bm{q}}_d^T]^T \in \mathbb{R}^8,
\end{equation}
where  \lx{the ``$+$'' operator in $\tilde{\bm{q}}_r + \varepsilon\tilde{\bm{q}}_d$ represents the formal dual-number composition of the two 4D quaternions}, $\tilde{\bm{q}}_r \in \mathbb{R}^4$ is the real part, $\tilde{\bm{q}}_d \in \mathbb{R}^4$ is the dual part, and $\varepsilon$ is Clifford's dual unit satisfying $\varepsilon \neq 0$ and $\varepsilon^2 = 0$. 

The multiplication $\otimes$ of $\hat{\bm{q}}_1 = \tilde{\bm{p}}_1 + \varepsilon\tilde{\bm{q}}_1$ and $\hat{\bm{q}}_2 = \tilde{\bm{p}}_2 + \varepsilon\tilde{\bm{q}}_2$ is given by:
\begin{equation}
\label{eq:dq_mult_def}
\hat{\bm{q}}_1 \otimes \hat{\bm{q}}_2 = \tilde{\bm{p}}_1 \odot \tilde{\bm{p}}_2 + \varepsilon\big(\tilde{\bm{p}}_1 \odot \tilde{\bm{q}}_2 + \tilde{\bm{q}}_1 \odot \tilde{\bm{p}}_2\big).
\end{equation}
Three conjugates are defined for $\hat{\bm{q}} = \tilde{\bm{q}}_r + \varepsilon\tilde{\bm{q}}_d$: $\hat{\bm{q}}^{1*} := \tilde{\bm{q}}_r - \varepsilon\tilde{\bm{q}}_d$, $\hat{\bm{q}}^{2*} := \tilde{\bm{q}}_r^* + \varepsilon\tilde{\bm{q}}_d^*$, and $\hat{\bm{q}}^{3*} := \tilde{\bm{q}}_r^* - \varepsilon\tilde{\bm{q}}_d^*$.

A rigid-body pose in $SE(3)$ with rotation $\tilde{\bm{q}}_r \in \mathcal{S}^3$ and translation $\bm{t} \in \mathbb{R}^3$ (embedded as $\tilde{\bm{t}} = [0, \, \bm{t}^T]^T$) is represented by a unit dual quaternion \eqref{eq:dq_def} with $\tilde{\bm{q}}_d = \frac{1}{2}\tilde{\bm{t}} \odot \tilde{\bm{q}}_r$, which satisfies $\|\tilde{\bm{q}}_r\|_2 = 1$, $\tilde{\bm{q}}_r^T\tilde{\bm{q}}_d = 0$, $\hat{\bm{q}}^{3*} \otimes \hat{\bm{q}}^{1*} = \tilde{\bm{1}} = [1, \, 0, \, 0, \, 0]^T$, and $\tilde{\bm{t}} = 2\tilde{\bm{q}}_d \odot \tilde{\bm{q}}_r^*$. Embedding $\bm{a}, \bm{b} \in \mathbb{R}^3$ as $\hat{\bm{a}} = \tilde{\bm{1}} + \varepsilon\tilde{\bm{a}}$ and $\hat{\bm{b}} = \tilde{\bm{1}} + \varepsilon\tilde{\bm{b}}$, the rigid-body transformation $\bm{a} = \bm{C}(\tilde{\bm{q}}_r)\bm{b} + \bm{t}$ takes the form $\hat{\bm{a}} = \hat{\bm{q}} \otimes \hat{\bm{b}} \otimes \hat{\bm{q}}^{3*}$. Right-multiplying both sides by $\hat{\bm{q}}^{1*}$ yields $\hat{\bm{a}} \otimes \hat{\bm{q}}^{1*} = \hat{\bm{q}} \otimes \hat{\bm{b}}$, whose dual part gives the fundamental single-point linear constraint \cite{srivatsan2016estimating, li2022pose}:
\begin{equation}
\label{eq:dq_dual_part_eq}
\tilde{\bm{a}} \odot \tilde{\bm{q}}_r - \tilde{\bm{q}}_r \odot \tilde{\bm{b}} - 2\tilde{\bm{q}}_d = \big[\bm{L}(\tilde{\bm{a}}) - \bm{R}(\tilde{\bm{b}})\big]\tilde{\bm{q}}_r - 2\tilde{\bm{q}}_d = \tilde{\bm{0}}.
\end{equation}

\section*{ACKNOWLEDGMENT}

The authors utilized Google Gemini to assist with English language editing, phrasing refinement, and minor code debugging during the preparation of this manuscript. No generative AI tools were used to generate scientific insights, theoretical derivations, or experimental data. After using this tool, the authors thoroughly reviewed, edited, and verified all text, mathematical formulations, and simulation codes, and take full responsibility for the integrity and accuracy of the final manuscript.

\renewcommand{\refname}{\normalfont\rmfamily\scshape\normalsize REFERENCES}
\bibliographystyle{IEEEtran}
\bibliography{references}

\end{document}